\documentclass[12pt,a4paper]{article}

\usepackage{amsmath}
\usepackage{amssymb}
\usepackage{graphicx}
\usepackage{multirow}
\usepackage{booktabs}
\usepackage{tabularx}
\usepackage{hyperref}
\usepackage{algorithm}
\usepackage{algpseudocode}
\usepackage{subcaption}
\usepackage{url}

\DeclareCaptionLabelFormat{withbar}{\textbf{#1~#2~\textbar~}}

\hypersetup{
    colorlinks=true,
    linkcolor=blue,    
    citecolor=blue,      
    urlcolor=blue,       
    bookmarksnumbered=true
}

\makeatletter
\renewcommand\@cite[1]{\textsuperscript{#1}}
\renewcommand\@biblabel[1]{#1.}
\makeatother

\begin{document}
\setlength{\parindent}{2em} 

\begin{center}
{\Large \textbf{MAFS: Multi-head Attention Feature Selection for High-Dimensional Data via Deep Fusion of Filter Methods}}

\vspace{0.5cm}

\textbf{Xiaoyan Sun}\textsuperscript{1,†}, \textbf{Qingyu Meng}\textsuperscript{1,†}, \textbf{Yalu Wen}\textsuperscript{1,*}

\vspace{0.3cm}

\textsuperscript{1}Department of Statistics, University of Auckland, Auckland, New Zealand

\vspace{0.2cm}
\textsuperscript{†}These authors contributed equally to this work.

\textsuperscript{*}To whom correspondence should be addressed: [y.wen@auckland.ac.nz]

\end{center}

\vspace{0.5cm}

\section*{Abstract}
\label{sec:abstract}

Feature selection is essential for high-dimensional biomedical data,  enabling stronger predictive performance, reduced computational cost, and improved interpretability in precision-medicine applications. Existing approaches face notable challenges. Filter methods are highly scalable but cannot capture complex relationships or eliminate redundancy. Deep learning-based approaches can model nonlinear patterns and but often lack stability, interpretability, and efficiency at scale. Single-head attention improves interpretability but is limited in capturing multi-level dependencies and remains sensitive to initialization, reducing reproducibility. Most existing methods rarely combine statistical interpretability with the representational power of deep learning, particularly in ultra–high-dimensional settings. Here, we introduce MAFS (Multi-head Attention–based Feature Selection), a hybrid framework that integrates statistical priors with deep learning capabilities. MAFS begins with filter-based priors for stable initialization and guide learning. It then uses multi-head attention to examine features from multiple perspectives in parallel, capturing complex nonlinear relationships and interactions. Finally, a reordering module consolidates outputs across attention heads, resolving conflicts and minimizing information loss to generate robust and consistent feature rankings. This design combines statistical guidance with deep modeling capacity, yielding interpretable importance scores while maximizing retention of informative signals. Across simulated and real-world datasets, including cancer gene expression and Alzheimer’s disease data, MAFS consistently achieves superior coverage and stability compared with existing filter-based and deep learning–based alternatives, offering a scalable, interpretable, and robust solution for feature selection in high-dimensional biomedical data.

\newpage
\section{Introduction}
\label{sec:intro}

Precision medicine aims to tailor prevention and treatment strategies to individual genetic, environmental, and lifestyle factors\cite{ashley2016towards}. The rapid expansion of high-throughput biotechnology and large-scale biobanks has enabled the collection of multi-omics data, encompassing genomics, transcriptomic, epigenomics and other molecular layers, alongside clinical and lifestyle information\cite{chen2023graph,uffelmann2021genome,motsinger2024gene}. These datasets are ultra high-dimensional, heterogeneous, and multimodal, capturing complex biological processes across multiple molecular scales. Extracting meaningful biological signals from such data requires efficient feature selection, which improves downstream model performance, reduce computational complexity, and enhance interpretability and reproducibility\cite{yang2021feature,wang2023scientific}, providing a foundation for precise diagnosis, drug development, and personalized treatment.

Feature selection is typically categorized into filter, wrapper, and embedded methods\cite{liu2019embedded,saeys2007review,guyon2003introduction}. For ultra high-dimensional data, wrapper methods are often computationally infeasible\cite{bommert2022benchmark,guyon2002gene}, while embedded methods (e.g., LASSO and tree-based models\cite{tibshirani1996regression, clark2017tree}) integrate feature selection into model training but remain limited in scalability for such settings. In contrast, filter methods independently evaluate feature relevance using statistical or correlation measures without requiring model training, offering high scalability and efficiency for preliminary screening. For example, Sure Independence Screening (SIS)\cite{fan2008sure} uses marginal correlations to efficiently identify promising features; its extensions, such as DC-SIS, incorporate distance correlation to capture nonlinear dependencies\cite{li2012feature}, while CSIS accounts for conditional contributions to reduce confounding\cite{barut2016conditional}. Robust alternatives, including Kendall rank correlation\cite{hastie2009introduction} and Ball correlation-based SIS (BCor-SIS)\cite{pan2019generic}, further improve resilience to non-normality and nonlinear effects. Overall, filter methods provide multi-perspective low-cost priors but cannot eliminate redundant features and remain constrained by their assumed feature-outcome relationship, which  may overlook key features and complex interactions\cite{guyon2003introduction}.  

Deep learning is naturally capable of capturing complex, nonlinear relationships, making it a powerful tool for feature selection\cite{huang2023evaluation}. Depending on architecture and mechanism, deep learning-based approaches include graph-based methods (e.g., GRACES) that leverage sample–feature relationships but may face scalability challenges\cite{chen2023graph}; gradient attribution methods (e.g., DeepLIFT and GradientShap) that offer computational efficiency and intuitive interpretation but sensitive to reference baselines and gradient noise\cite{shrikumar2017learning,lundberg2017unified}; perturbation-based methods (e.g., LIME and FeatureAblation) that are model-agnostic and can measure local importance but computationally demanding and sensitive to sampling strategies\cite{ribeiro2016should,kokhlikyan2020captum}; architectural layer methods (e.g., CancelOut and autoencoders) that learn compact representations during training but may suffer from instability and sensitivity to regularization or initialization\cite{borisov2019cancelout,bietti2019group,achille2018emergence}. 

Among deep learning approaches, attention mechanisms are particularly well-suited for feature selection due to their ability to learn importance weights without requiring explicit feature engineering or domain-specific assumptions\cite{vaswani2017attention}. Originally developed for neural machine translation\cite{bahdanau2014neural} and later advanced in the Transformer architecture\cite{vaswani2017attention}, attention mechanisms can selectively focus on the most relevant features. External attention mechanisms\cite{guo2022beyond} further improve computational efficiency by reducing complexity to linear scale using shared memory units, making them highly applicable for feature selection. For example, Gui \textit{et al.} first proposed the AFS\cite{gui2019afs}, a supervised feature selection method based on attention mechanisms. EAR-FS\cite{xue2023external} further advanced this approach by integrating external attention mechanisms into feature selection, demonstrating superior capability in identifying relevant features from high-dimensional genomic data.  Despite these advantages, traditional single attention mechanisms typically capture feature relationships along a single dimension, limiting their ability to fully represent complex interactions and multi-level dependencies in biological data. Additionally, like other deep learning approaches, attention mechanisms face challenges in stability, including sensitivity to initialization and variability in feature importance rankings across identical datasets. These issues are particularly critical for ensuring reliable biomarker discovery and robust feature selection in high-dimensional biomedical applications. 

To address these challenges, we introduce MAFS (Multi-head Attention-based Feature Selection), a novel framework that combines filter methods with multi-head attention mechanisms and identifies important features through a reordering module. MAFS begins with filter-based screening to generate initial feature priors, mitigating sensitivity to initialization and accelerating convergence. It then applies multi-head attention to process features from multiple perspectives in parallel, overcoming the limitations of single-attention mechanisms. Finally, a reordering module is designed to refine feature sets from each attention head, minimizing information loss and resolving conflicts. This deep fusion design combines the stability and efficiency of filter methods with the representational power of deep learning, offering interpretable feature importance scores while maximizing retention of critical features through prior knowledge guidance and enhanced reordering selection strategies.
\section{Results}
\label{sec:results}

\subsection{Methods overview and simulations}

\begin{figure}[htbp]
\centering
\begin{subfigure}{\textwidth}
    \includegraphics[width=\textwidth]{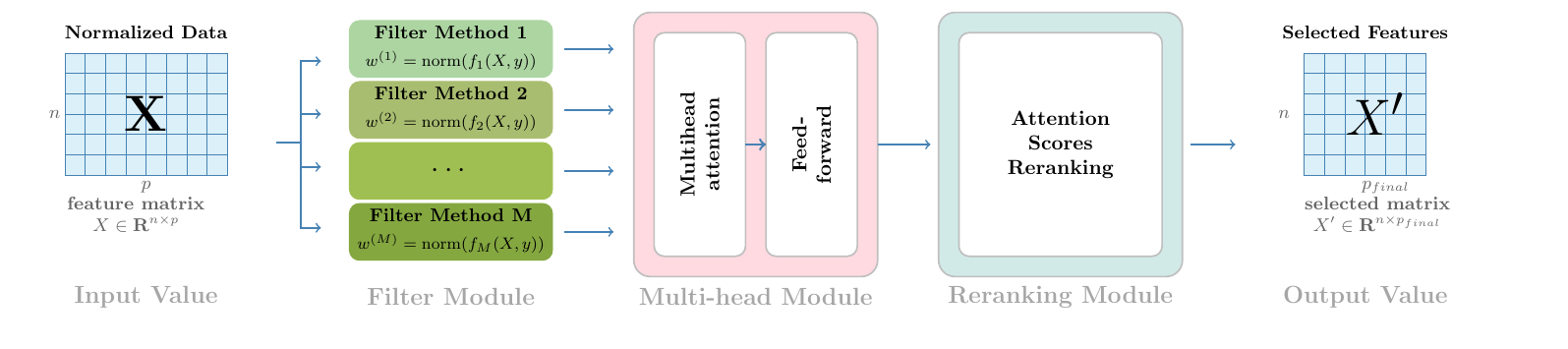}
    \subcaption{Overall architecture of MAFS}
    \label{fig:overall}
\end{subfigure}
\begin{subfigure}{\textwidth}
    \includegraphics[width=\textwidth]{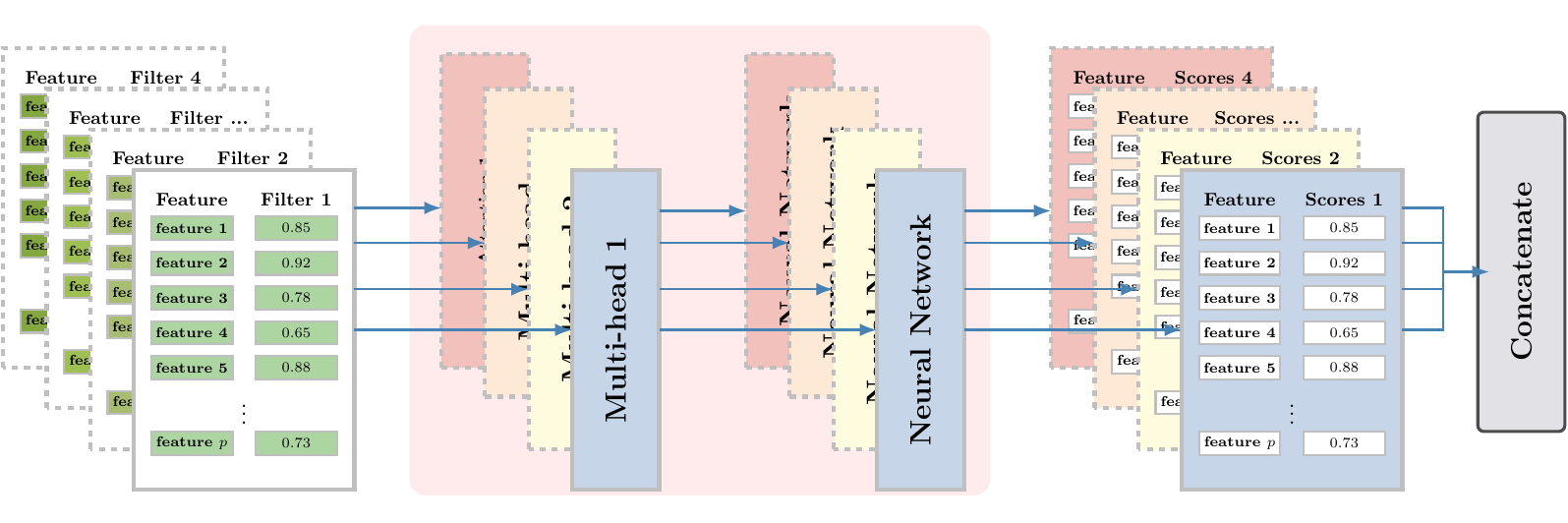}
    \subcaption{Multi-head attention module}
    \label{fig:attention}
\end{subfigure}
\begin{subfigure}{\textwidth}
    \includegraphics[width=\textwidth]{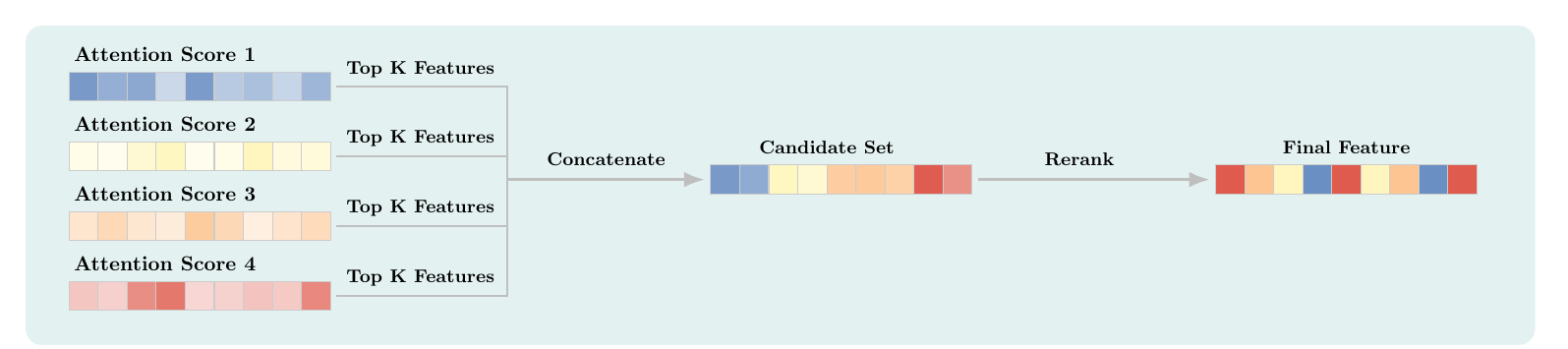}
    \subcaption{Reordering module}
    \label{fig:rerank}
\end{subfigure}

\caption{\textbf{Overview of MAFS.} (a) Overall architecture of the proposed Multi-head Attention-based Feature Selection (MAFS) framework, showing three main modules: Filter Module, Multi-head Attention Module, and Reordering Module. (b) Multi-head attention module where multiple attention heads process the input in parallel. Each head has its own set of learnable parameters to capture diverse feature patterns and interactions. (c) Reordering module that consolidates outputs from individual attention heads to identify the most informative features.}
\label{fig:method}

\end{figure}

We developed MAFS (Fig.~\ref{fig:overall}), a deep learning framework that combines statistical filtering with attention-based representation learning to achieve efficient, scalable, and interpretable feature selection in ultra-high-dimensional biomedical data. MAFS operates in three key steps. First, it applies multiple classical filter methods to generate preliminary relevance estimates, capturing feature importance under specific feature–outcome assumptions. Second, these filter-derived estimates guide a multi-head external attention mechanism, allowing complementary evaluation of feature importance from diverse perspectives (Fig.~\ref{fig:attention}). Finally, outputs from individual attention heads are consolidated through a re-ranking strategy to identify the most informative features (Fig.~\ref{fig:rerank}). Our design substantially improves both feature selection accuracy and computational efficiency compared with existing deep learning-based approaches.

\subsection{Relationship-specific feature selection performance}
\begin{figure}[H]
    \centering
    \includegraphics[width=0.95\textwidth]{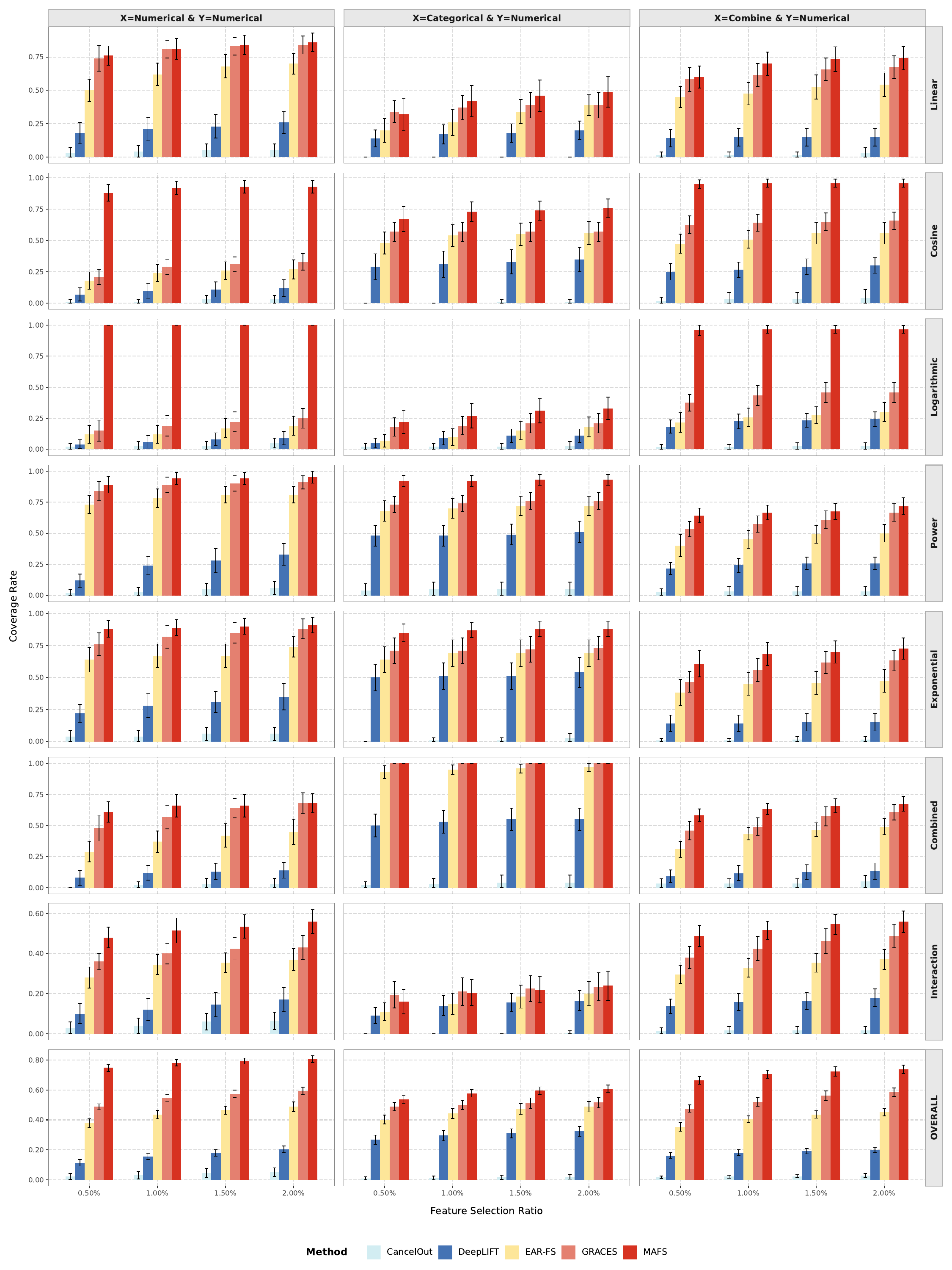}
    \caption{\textbf{Results of relationship-specific feature selection for continuous outcomes.} Coverage rates under seven different feature–response relationships at selection ratios 0.5\%, 1\%, 1.5\% and 2\% in the high-dimensional setting ($n=2{,}000$, $p=100{,}000$) for continuous outcomes.}
    \label{fig:relationship_2k}
\end{figure}

\begin{figure}[H] 
    \centering
    \includegraphics[width=0.95\textwidth]{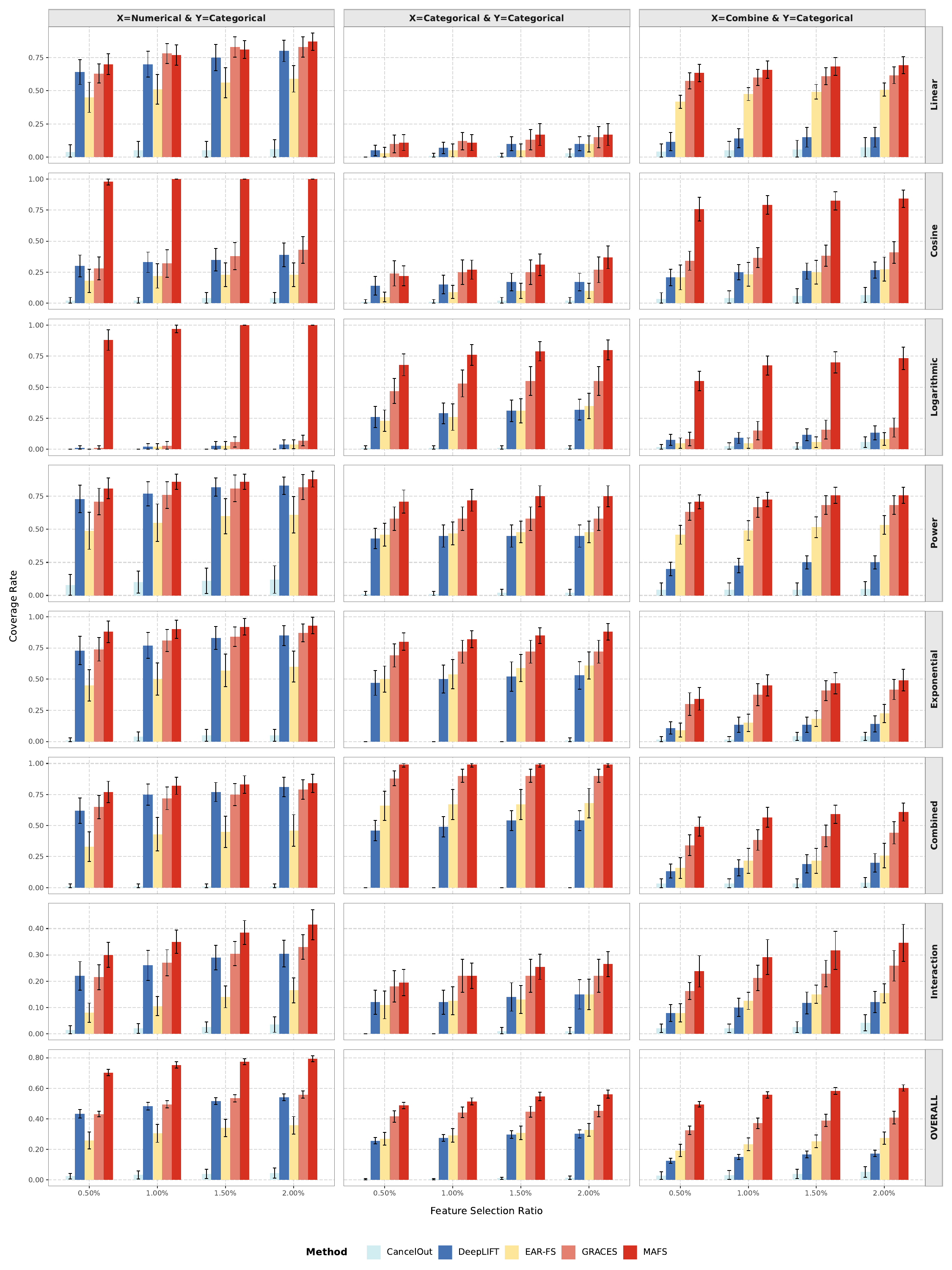}
    \caption{\textbf{Results of relationship-specific feature selection for binary outcomes.} Coverage rates under seven different feature–response relationships at selection ratios 0.5\%, 1\%, 1.5\% and 2\% in the high-dimensional setting ($n=2{,}000$, $p=100{,}000$) for binary outcomes.}
    \label{fig:relationship_analysis_2k_cate}
\end{figure}

Relationship-specific feature selection performance as the pre-specified feature selection ratio increased from 0.5\% to 2\% for high-dimensional and moderate sample size ($n=2,000$, $p=100,000$) is shown in Fig.~\ref{fig:relationship_2k} and Fig.~\ref{fig:relationship_analysis_2k_cate} for continuous and binary outcomes, respectively. Additional results for different sample sizes ($n=500$ and $n=2,000$), dimensions ( $p=25,000$, $p=50,000$ and $p=100,000$), and outcome types (continuous and binary) are provided in the Supplementary Fig. S1 - S10. Across all scenarios, MAFS consistently demonstrated superior performance among all methods. Additionally, MAFS displayed narrower confidence intervals ($\pm$2.2\%) compared to other methods ($\pm$2.5--3.1\%), demonstrating greater stability across replications.

When the pre-specified feature selection proportion increased from 0.5\% to 2\%, the number of selected features increased accordingly, leading to higher coverage rates across all methods and functional forms, as expected (Fig.~\ref{fig:relationship_2k} and \ref{fig:relationship_analysis_2k_cate}). However, the magnitude of improvement varied among methods, reflecting their differing abilities to prioritize informative features early in the selection process. For example, for continuous outcomes, although MAFS achieved the highest coverage across all pre-specified selection ratios, its coverage increased by only about 7.85\% on average as the selection proportion rose from 0.5\% to 2\%. This suggests that MAFS effectively identified the most informative features early, with limited additional benefit gained from expanding the feature set. In contrast, GRACES, EAR-FS and DeepLIFT showed improvements of 21.8\%, 29.1\%, and 79.1\%, respectively, suggesting that these methods required larger feature subsets to recover additional causal features and were less effective at early feature prioritization compared with MAFS. For CancelOut, although the improvements were substantial, its coverage rates consistently remained below 0.06, reflecting its limited effectiveness in this setting.

The relative performance of feature selection methods varied substantially across different functional relationships between features and outcomes. Using continuous features and continuous outcomes as an example, linear settings produced broadly comparable performance among the top methods: MAFS achieved the highest coverage (0.86), followed closely by GRACES (0.84) and EAR-FS (0.70), whereas DeepLIFT (0.26) and CancelOut (0.05) performed substantially worse. As the underlying relationships became more nonlinear, performance gaps widened sharply. For logarithmic relationships, MAFS reached near-perfect coverage (1.0), far exceeding GRACES (0.25), EAR-FS (0.19), DeepLIFT (0.09), and CancelOut(0.05). A similar pattern appeared in cosine relationships, where MAFS achieved 0.93 coverage compared with 0.33 and 0.27 for the strongest competing methods, GRACES and EAR-FS, respectively. Even for moderately complex relationships, such as exponential (0.91 vs. 0.88 vs. 0.74), power (0.95 vs. 0.91 vs. 0.81), and interaction effects (0.56 vs. 0.43 vs. 0.37), MAFS consistently outperformed both GRACES and EAR-FS. These results highlight that although strong baselines perform competitively in simple settings, MAFS delivers substantially greater robustness as functional complexity increases.

We further observed that the complexity of feature–outcome relationships influenced the benefits of increasing the feature selection depth. For relationships such as logarithmic or cosine functions, expanding the number of selected features did not substantially increase coverage for MAFS, which achieved near-perfect coverage ($\geq$0.80) even at 0.5\% selection. In contrast, for more complex relationships such as interactions, MAFS showed greater gains from expanded feature subsets, with coverage improving from 0.48 to 0.56 as the selection proportion increased from 0.5\% to 2\%. Other relationships exhibited intermediate patterns, with moderate improvements ranging from 0.05 to 0.12. These findings indicate that larger selected feature subsets may be beneficial when features and outcomes are expected to have more complex functional dependencies. 

The performance of feature selection methods was further influenced by feature and outcome variable types. When examining continuous outcomes, methods generally achieved higher coverage with continuous features compared to categorical features. For continuous features, MAFS achieved 0.806 coverage, substantially outperforming GRACES (0.594), EAR-FS (0.488), DeepLIFT (0.204), and CancelOut (0.051). When features were categorical, the performance gaps between methods narrowed. For instance, MAFS achieved 0.653 coverage with categorical features, while GRACES and EAR-FS reached 0.487 and 0.432, respectively, demonstrating reduced relative advantages compared to the continuous feature setting. Although GRACES was specifically designed for small-sample settings, MAFS demonstrated comparable or superior performance even under these conditions (Supplementary Fig. S1 - S6). Results under other dimensional settings and smaller sample sizes remained consistent, preserving similar relative performance patterns across different combinations of feature and outcome types.

\subsection{The Impact of Feature Dimensionality}
\subsubsection{The Impact of Feature Dimensionality: Feature Selection Accuracy}

\begin{figure}[H] 
    \centering
    \includegraphics[width=\textwidth]{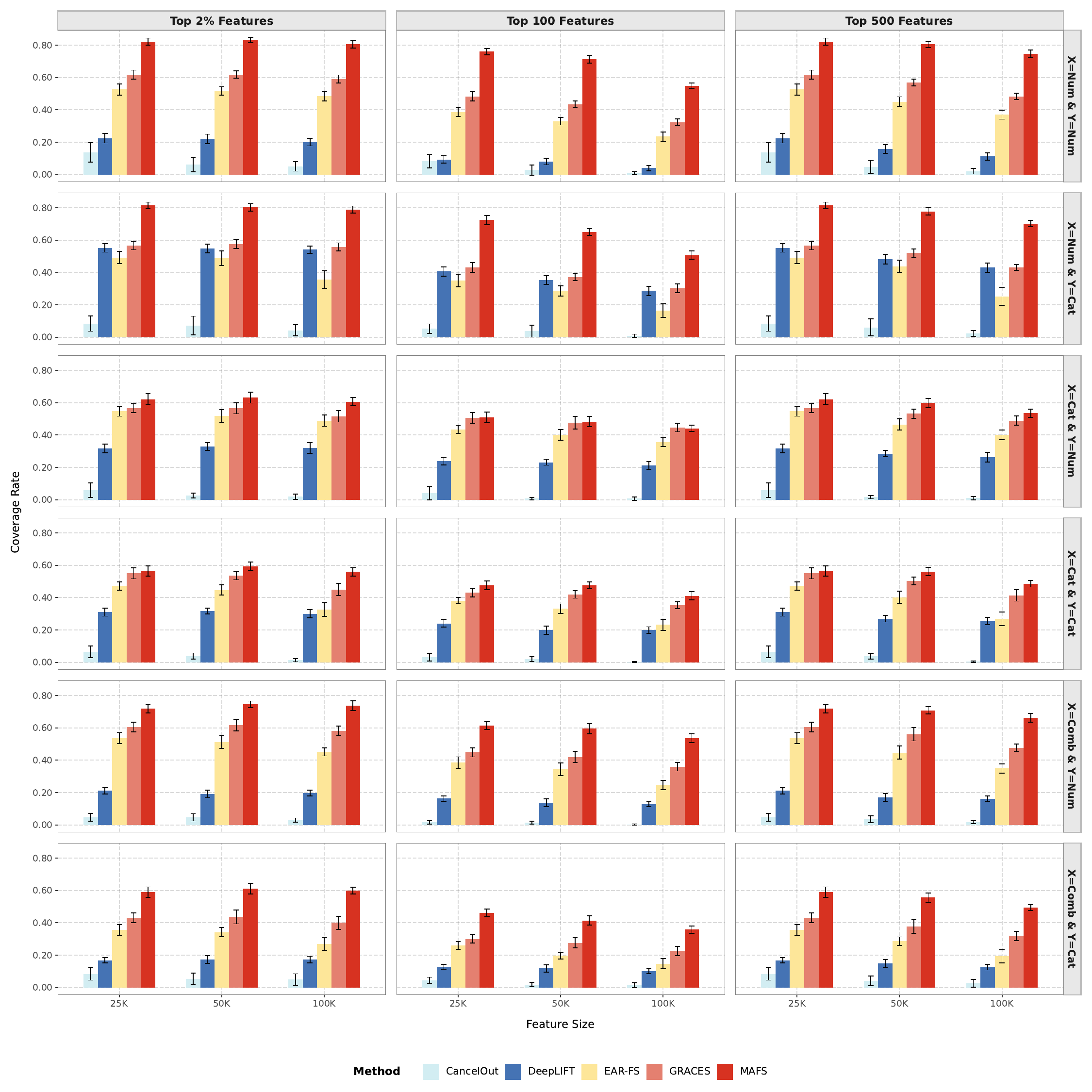}
    \caption{\textbf{Feature selection performance across different dimensionalities.} Coverage rates under different input feature dimensions with moderate sample size ($n=2{,}000$). Each subfigure compares five feature selection methods across three feature dimensionalities (25K, 50K, and 100K features). Rows correspond to six data type combinations defined by feature types (continuous, categorical, and combined) and response types (continuous and binary). Columns correspond to three selection criteria: top 2\%, top 100, and top 500 features.}
    \label{fig:dimension_2k}
\end{figure}

Fig.~\ref{fig:dimension_2k} and Supplementary Fig. S11 present the performance of the feature selection methods as the number of input features increases from 25,000 to 100,000 for $n=2,000$ and $n=500$, respectively. Across all input dimensions, the proposed MAFS consistently achieves the highest coverage, regardless of sample size, feature type, or outcome type. In the majority of settings, its performance is followed by GRACES, EAR-FS, DeepLIFT, and CancelOut. An exception arises when the features are continuous and the outcome is binary, wherein DeepLIFT exceeds the performance of EAR-FS, resulting in the ordering MAFS, GRACES, DeepLIFT, EAR-FS, and CancelOut.

When 2\% of the top-ranked features were selected, the numbers of selected features were 500, 1,000, and 2,000 for input dimensions of 25,000, 50,000, and 100,000, respectively. As shown in both Fig.~\ref{fig:dimension_2k}A and Supplementary Fig. S11.A, the coverage rates for most methods maintained stable or showed slight improvements as dimensionality increased. This stability likely arose because the number of selected features grew proportionally with dimensionality, effectively mitigating the curse of dimensionality. 

In contrast, when a fixed number of top-ranked features was selected (i.e., top 100 and top 500), all methods exhibited the expected decline in coverage, with MAFS remaining the top performer and showing the smallest percentage reduction in coverage rate. When restricting selection to the top 100 features, coverage rates decreased across methods as input dimensionality increased. For example, when both the features and outcomes were continuous with a sample size of 2,000, the coverage rate for MAFS declined from 0.6 to 0.475 as the number of input features increased from 25,000 to 100,000, corresponding to a 21\% reduction in the coverage rate. For GRACES (0.375 to 0.20), EAR-FS (0.325 to 0.1), DeepLIFT (0.125 to 0.025), and CancelOut (0.075 to 0), the corresponding percentage reduction were 47\%, 69\%, 80\%, 100\%, respectively. A similar pattern was observed when the features were continuous and the outcome was binary. Under the same settings, the coverage rate decreased by 38\%, 44\%, 87\%, 50\%, and 100\% for MAFS (0.575 to 0.375), GRACES (0.40 to 0.225), EAR-FS (0.375 to 0.05), DEEPLift (0.40 to 0.20) and CancelOut (0.10 to 0), respectively. Although some methods (e.g., CancelOut) may exhibit relatively small absolute changes in coverage, they often experience large proportional reductions and maintain overall coverage levels that are extremely low (e.g., below 5\%), rendering them substantially less suitable for dimensionality reduction in high-dimensional settings. When the top 500 features were selected, similar declining patterns were observed, although the reductions were more moderate as input dimension increases. 

These simulation results suggest that increasing dimensionality poses substantial challenges for all feature selection methods. Although MAFS consistently exhibited the best performance across all input dimensions considered, selecting a slightly larger subset of features in high-dimensional settings can help recover more outcome-relevant variables, likely benefiting downstream tasks.

\subsubsection{The Impact of Feature Dimensionality: Computational Efficiency}

Table \ref{tab:runtime_minutes} presents the runtime distributions of the five feature selection methods across different sample sizes ($n=500$ and $n=2{,}000$) and feature dimensions ($p=25{,}000$, $50{,}000$, and $100{,}000$). All experiments were conducted on the REANNZ high-performance computing platform\cite{reannz}, and runtime is reported in minutes.

Runtime differed slightly between continuous and binary outcomes. MAFS required more computation time for continuous outcomes, and this gap widened as the dimensionality increased. For example, at $p=100{,}000$ and $n=2{,}000$, MAFS required 117.6 min for continuous outcomes compared with 88.5 min for binary outcomes. The remaining  methods exhibited only marginal differences across outcome types.

In high-dimensional settings ($n=2{,}000$, $p=100{,}000$), runtime varied substantially across methods. For continuous outcomes, CancelOut, DeepLIFT, and EAR-FS remained computationally efficient (20.2, 13.2, and 29.1 min, respectively), whereas MAFS required 117.6 min. GRACES, the strongest baseline comparator, incurred a markedly higher computational cost of 1,569.9 min. A similar pattern was observed for binary outcomes, where runtimes were 25.4 min (CancelOut), 14.3 min (DeepLIFT), 28.4 min (EAR-FS), 88.5 min (MAFS), and 1,540.0 min (GRACES).

Although MAFS was slower than the fastest lightweight baselines, it achieves a strong balance between computational efficiency and selection performance. Importantly, MAFS is more than an order of magnitude faster than GRACES, the best-performing baseline method, while simultaneously offering substantially higher coverage across all evaluated scenarios.

\begin{table}[htbp]
\centering
\resizebox{\textwidth}{!}{
\begin{tabular}{llccc|ccc}
\toprule[1.5pt]
\multirow{3}{*}{Outcome} & \multirow{3}{*}{Method} 
& \multicolumn{6}{c}{Sample size} \\
\cmidrule(lr){3-8}
& & \multicolumn{3}{c|}{$n=500$} & \multicolumn{3}{c}{$n=2000$} \\
\cmidrule(lr){3-5} \cmidrule(lr){6-8}
& & 25k & 50k & 100k & 25k & 50k & 100k \\
\midrule
\multirow{5}{*}{continuous} 
 & MAFS & 5.9 [5.6, 6.2] & 11.3 [10.9, 11.7] & 44.2 [38.4, 50.0] 
         & 16.7 [15.5, 18.0] & 47.8 [42.7, 53.0] & 117.6 [110.1, 125.0] \\
 & GRACES    & 83.0 [74.6, 91.4] & 176.1 [161.5, 190.6] & 380.2 [348.5, 412.0]
             & 253.4 [234.6, 272.1] & 601.3 [567.8, 634.9] & 1569.9 [1492.4, 1647.4] \\
 & EAR-FS    & 1.0 [1.0, 1.1] & 2.0 [1.8, 2.1] & 4.4 [3.5, 5.4]
             & 1.8 [1.5, 2.0] & 4.2 [3.6, 4.7] & 29.1 [23.7, 34.6] \\
 & DeepLIFT  & 0.5 [0.4, 0.6] & 3.4 [2.6, 4.2] & 4.2 [3.2, 5.2]
             & 1.3 [1.1, 1.4] & 7.0 [5.5, 8.5] & 13.2 [11.4, 15.1] \\
 & CancelOut & 0.6 [0.5, 0.7] & 1.3 [1.1, 1.5] & 4.1 [3.4, 4.8]
             & 2.3 [2.0, 2.6] & 3.9 [3.4, 4.3] & 20.2 [15.0, 25.4] \\
\midrule
\multirow{5}{*}{binary} 
 & MAFS & 5.5 [5.2, 5.8] & 11.2 [9.7, 12.7] & 40.6 [33.7, 47.6]
        & 11.3 [9.9, 12.7] & 26.4 [23.1, 29.8] & 88.5 [81.8, 95.2] \\
 & GRACES    & 86.9 [79.6, 94.3] & 169.8 [130.6, 209.1] & 392.8 [374.0, 411.6]
             & 285.7 [260.6, 310.8] & 555.7 [522.4, 589.0] & 1540.0 [1421.6, 1658.4] \\
 & EAR-FS    & 1.0 [1.0, 1.1] & 2.0 [1.8, 2.1] & 6.1 [5.0, 7.1]
             & 1.7 [1.5, 1.8] & 4.7 [4.2, 5.2] & 28.4 [24.7, 32.1] \\
 & DeepLIFT  & 0.6 [0.5, 0.7] & 2.9 [2.3, 3.5] & 6.4 [4.9, 7.8]
             & 1.6 [1.5, 1.8] & 7.9 [6.5, 9.3] & 14.3 [12.5, 16.1] \\
 & CancelOut & 0.8 [0.6, 0.9] & 1.4 [1.2, 1.7] & 4.4 [3.5, 5.3]
             & 1.9 [1.6, 2.1] & 3.5 [3.0, 4.0] & 25.4 [22.7, 28.1] \\
\bottomrule[1.5pt]
\end{tabular}
}
\caption{Runtime (in minutes) of five feature selection methods across different sample sizes and feature dimensions, reported with 95\% confidence intervals.}
\label{tab:runtime_minutes}
\end{table}

\subsection{Real data analysis}
\subsubsection{The Analysis of Cancer Gene Expression Datasets}

\begin{figure}[H] 
    \centering
    \includegraphics[width=0.95\textwidth]{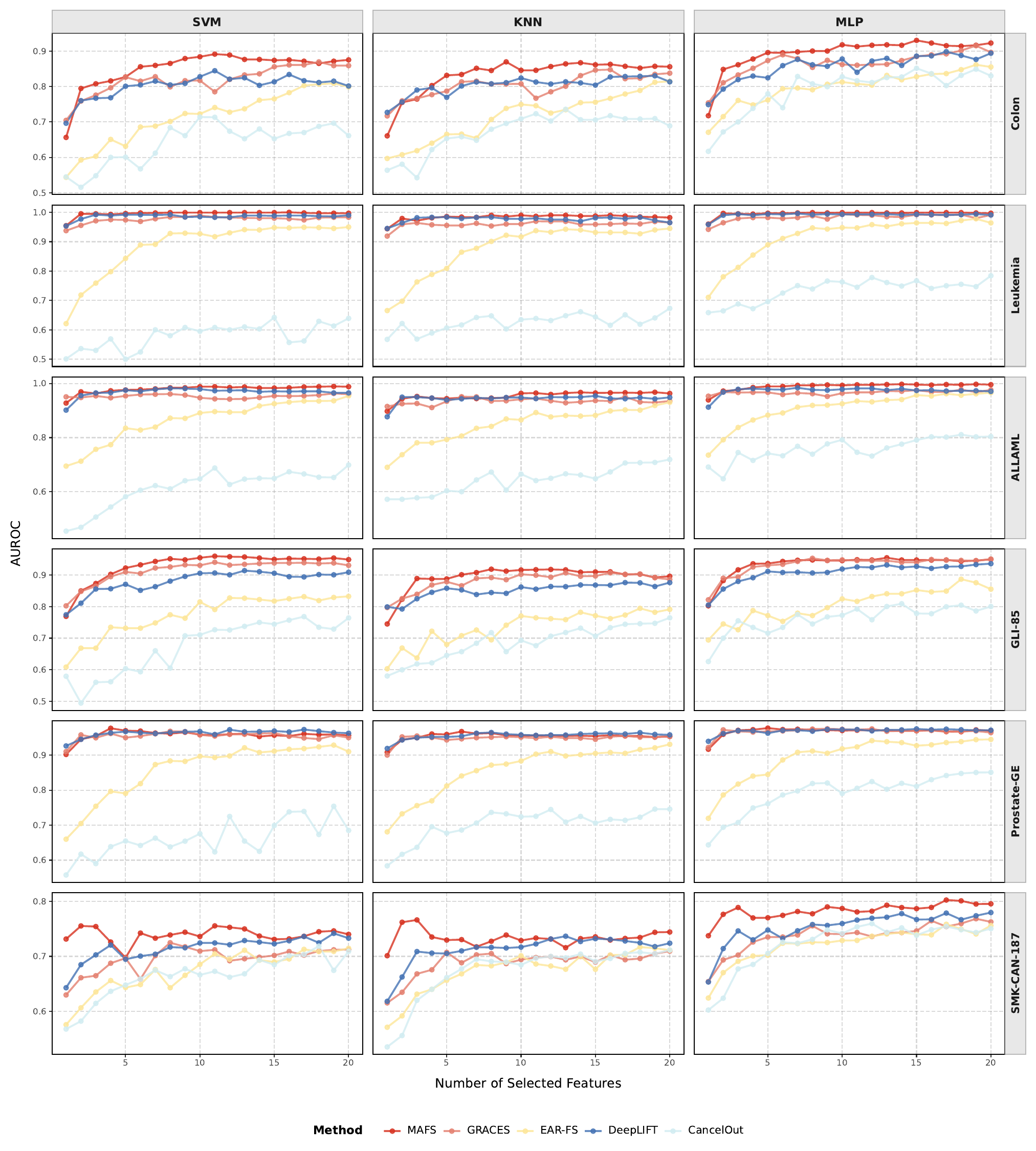}
    \caption{\textbf{Evaluation on Cancer Gene Expression Datasets.} Comparison of AUROC performance for feature selection methods across six cancer datasets. Each row corresponds to one dataset, and each column represents a different classifier: Support Vector Machine (left), K-Nearest Neighbors (middle), and Multi-Layer Perceptron (right).}
    \label{fig:auroc_all_classifiers}
\end{figure}

Fig.~\ref{fig:auroc_all_classifiers} shows the AUROC of all methods across the six cancer datasets. Overall, classification performance was highest for the MLP, followed by SVM and KNN, although differences among the three classifiers were small. Importantly, the relative performance of feature selection methods varied across datasets but was consistent across classifiers. On easily separable datasets (i.e., leukemia and prostate cancer), most methods achieved high performance (AUROC $\geq 0.95$). On the moderately difficult dataset (glioma), MAFS outperformed all baseline methods under SVM and KNN, but showed comparable performance to GRACES under MLP (AUROC $\approx$ 0.95). On the more challenging datasets (colon cancer and lung cancer), MAFS markedly outperformed all baseline methods. Under MLP with 20 features selected, it achieved an AUROC of 0.922 on colon cancer, exceeding GRACES by 0.026 and EAR-FS by 0.067, and reached 0.785 on lung cancer, surpassing GRACES by 0.022 and other methods by over 0.030.

MAFS maintained consistent relative performance across all three classifiers. Although absolute performance varied among classifiers, the relative ranking of methods remained stable. Across the 18 test scenarios combining 3 classifiers and 6 cancer datasets, MAFS achieved the highest or tied-for-highest performance in 17 of 18 scenarios. In the single exception (prostate cancer data set with SVM as the classifier), MAFS showed comparable performance to DeepLIFT while still outperforming all remaining comparison methods. This consistency across classifiers suggests that performance differences reflect feature selection effectiveness rather than sensitivity to specific classifier architectures.

MAFS also demonstrated efficient feature utilization, suggesting that MAFS converges more rapidly to optimal feature subsets compared to baseline methods. For instance, on the leukemia dataset, MAFS required only 2-3 features to achieve AUROC $>0.95$, whereas GRACES needed 4-6 features and EAR-FS required 8-10 features to reach comparable performance. On the colon dataset, MAFS attained an AUROC of 0.90 with 5 features, while baseline methods required 10-15 features. Similar patterns were observed  across other datasets, indicating that MAFS can achieve high performance with smaller feature subsets, which may reduce computational requirements and enhance interpretability in downstream analyses.

\subsubsection{The analysis of Alzheimer's Disease Neuroimaging Initiative Dataset}

\begin{figure}[H] 
    \centering
    \includegraphics[width=0.85\textwidth]{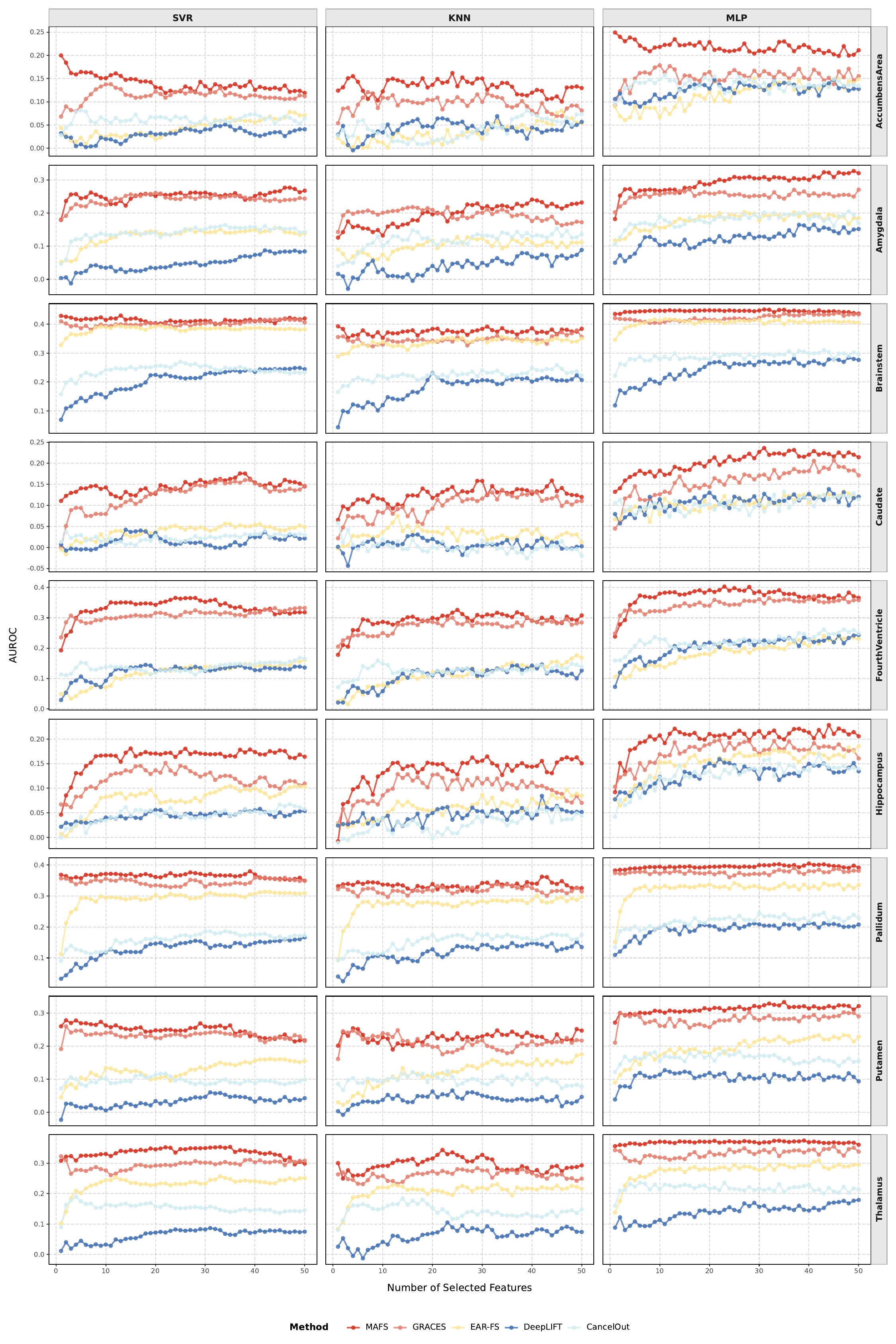}
    \caption{\textbf{Evaluation on ADNI Datasets.} Comparison of correlation performance for feature selection methods across the ADNI dataset. Each row corresponds to one brain region, and each column represents results obtained using a different regression model: Support Vector Regression (left), K-Nearest Neighbors Regression (middle), and Multi-Layer Perceptron Regression (right).}
    \label{fig:correlation_all}
\end{figure}

Fig.~\ref{fig:correlation_all} shows the Pearson correlation coefficients of all feature selection methods across nine image-derived phenotypes. Overall, correlation values varied across brain regions, suggesting that gene expression exerts region-specific contributions. MAFS generally achieved higher or comparable correlations than baseline methods across most regions and regressors, indicating effective selection of predictive features for continuous phenotype prediction.

Correlation values differed across brain regions. With 50 features selected, higher correlations were observed in the brainstem and pallidum, where MAFS achieved 0.440 and 0.395 (MLP), respectively, compared with GRACES (0.418 and 0.375), EAR-FS (0.409 and 0.329), DeepLIFT (0.228 and 0.190), and CancelOut (0.288 and 0.222). Moderate correlations were observed in amygdala, fourth ventricle, putamen, and thalamus, with MAFS consistently matching or slightly exceeding baseline methods. For example, in the fourth ventricle, MAFS achieved a correlation of 0.372, compared with 0.344 for GRACES, 0.192 for DeepLIFT, 0.175 for EAR-FS, and 0.214 for CancelOut. Lower correlations were found in the accumbens area, caudate, and hippocampus, where MAFS still reached the highest or comparable values among methods (e.g., 0.219 in accumbens area vs. 0.158 for GRACES, 0.125 for DeepLIFT, 0.100 for EAR-FS, and 0.131 for CancelOut).

MAFS maintained stable relative performance across regressors. While absolute correlation values varied, the ranking of methods was largely consistent. Across the 27 test scenarios (9 regions × 3 regressors), MAFS ranked first in the majority of scenarios, with competitive performance in the remaining cases, suggesting that its performance advantage arises from effective feature selection rather than sensitivity to specific regressor architectures. MAFS also demonstrated efficient feature utilisation, reaching high correlation values with fewer features than baseline methods. This efficient feature usage may reduce computational cost and facilitate interpretation in downstream neuroimaging analyses.

\section{Discussion}
\label{sec:discussion}

In this study, we introduced MAFS, a unified feature selection framework designed to address three key challenges in feature selection for high-dimensional biomedical data: 1) the irreversible signal loss during preliminary screening in conventional filter–wrapper pipelines, 2) the narrow assessment of feature relevance in single-criterion approaches, and 3) the initialization sensitivity that often destabilizes deep learning–based methods. MAFS addresses these issues by integrating multiple filter methods into a multi-head attention architecture, transforming statistical relevance scores into soft, learnable priors for each attention head. These priors guide the optimization process without acting as hard thresholds, remaining adjustable through backpropagation. This design preserves the computational efficiency and interpretability of filter methods while leveraging the representational power and multi-perspective modeling capabilities of attention mechanisms to capture complex dependencies in high-dimensional molecular data. Unlike existing embedding approaches that aggregate information from multiple perspectives via majority voting, we designed a reordering module to capture important features, even if they are identified by only a subset of perspectives. Through extensive simulations and analyses of multiple real datasets with both categorical and continuous outcomes, we demonstrate that MAFS achieves superior or comparable performance to the best-performing commonly used deep learning–based feature selection methods.

The simulation experiments revealed two notable trends. First, although all methods experienced declining coverage rates as feature dimensionality increased (Fig.~\ref{fig:dimension_2k}), MAFS maintained high accuracy even in the most challenging settings. This robustness is largely attributable to informed initialization, which provides a more favorable optimization landscape compared to random or uniform starting values. Informed initialization positions the model closer to optimal parameter regions, accelerating convergence and helping the attention mechanism focus on informative features from early training stages\cite{glorot2010understanding}. Second, across seven feature–outcome relationships, MAFS achieved the highest coverage rates in nearly all configurations, with particularly pronounced advantages for log-transformed and other nonlinear dependencies (Fig.~\ref{fig:relationship_2k} and \ref{fig:relationship_analysis_2k_cate}). The multi-head architecture enables complementary statistical perspectives to operate in parallel, mitigating the limitations of any single criterion. Heads initialized with different filter methods can capture common and unique  feature-outcome dependencies, reducing the risk of missing important features that would be overlooked by any single criterion. Coupled with the inherent capacity of deep neural networks to learn nonlinear and high-order effects, MAFS is well positioned for biomedical applications where data are high dimensional, sample sizes are limited\cite{chen2023graph}, and biological relationships\cite{reay2021advancing} are complex and heterogeneous. Although multi-head attention slightly increases computational overhead (Table~\ref{tab:runtime_minutes}), MAFS remains substantially faster than the best performing deep learning–based feature selection method (i.e., graph-based GRACES), offering a favorable balance between accuracy and efficiency.

Applications to real datasets further highlight MAFS’s practical value and its advantages over existing approaches. In cancer gene-expression datasets, MAFS matched or outperformed strong baselines such as GRACES and DeepLIFT, demonstrating robustness in high-dimensional, low-sample-size settings. Performance differences were minimal on easily separable datasets (leukemia and prostate cancer datasets), likely because strong feature–outcome associations allow most methods to identify discriminative signals, reducing the impact of fine-grained ranking. In contrast, on more challenging datasets (i.e., colon cancer and lung cancer), MAFS's multi-perspective learning enabled through informed initialization and multi-head attention facilitates more effective identification of informative features with diverse relationships, clearly benefiting downstream classification performance. A similar pattern was observed in the analysis of neuroimaging outcomes from ADNI, where signals were weaker and prediction tasks more difficult. Selecting features predictive of continuous cognitive scores requires capturing subtle, graded associations,  and the presence of higher noise amplifies the consequences of suboptimal feature ranking. Across brain regions, MAFS consistently outperformed all baselines, underscoring the contribution of its three core modules, including guided initialization, multi-head attention, and reordering, in producing stable and biologically meaningful feature rankings under weak-signal regimes.

It is worth noting that our proposed filter-weight–guided initialization alone can enhance attention based feature selection performance. This is consistent with previous findings showing that better initial values for deep networks can both improve performance and accelerate convergence toward optimal solutions\cite{glorot2010understanding}. Replacing uniform initialization in EAR-FS with filter-derived values consistently improved performance across simulated data (Supplementary Fig. S12 - S17), cancer gene-expression datasets (Supplementary Fig. S18), and ADNI neuroimaging data (Supplementary Fig. S19). While multi-head attention yielded additional gains, the fact that such a simple change in initialization, without modifying architecture or training protocol, consistently improves performance suggests that initialization quality may be as critical as architectural complexity for attention-based methods. These findings highlight that principled initialization is a transferable, low-cost strategy for strengthen existing attention-based frameworks without the need for a full architectural redesign.

While MAFS demonstrated clear advantages over existing methods, several limitations warrant consideration. First, although normalization is applied to ensure weights from different filters are on a comparable scale, simple normalization may still introduce scale misalignment and ranking instability. Exploring alternative fusion schemes, such as learnable scale parameters or rank-based aggregation, may help address this issue. Second, MAFS requires complete feature matrices, potentially excluding samples with missing values and introducing selection bias in biomedical datasets. Incorporating imputation or missing-data-aware mechanisms represents an important direction for future work.

In summary, MAFS integrates traditional filter methods with deep learning to tackle feature selection challenges in high-dimensional biomedical data. Through informed initialization,  multi-head attention and effecitve reordering, it achieves superior feature prioritization and robust performance across diverse data types, functional relationships, and dimensionality settings. 


\section{Methods}
\label{sec:Methods}

\subsection{The Models}

The proposed MAFS model addresses key challenges in feature selection for ultra-high dimensional biological data through a progressive architectural design comprising three core components (Fig.~\ref{fig:method}). The filter module employs multiple filter-based feature selection methods to generate an initial assessment of feature importance, each based on its own assumed feature–outcome relationship. The resulting weights provide prior knowledge for downstream learning, thereby enhancing model stability and convergence. The multi-head attention core module performs multi-perspective feature evaluation, where each attention head specializes in processing the output of one filter method, simultaneously evaluating feature importance from complementary viewpoints. The resulting attention weights enhance interpretability and transparency, enabling clearer understanding of feature importance and selection rationale. Additionally, the multi-head module adopts an external attention in place of self-attention, reducing computational complexity from $O(n^2p)$ to $O(np)$, thereby achieving substantial efficiency gains. The reordering module introduces an improved refinement that reorders and consolidates feature sets across attention heads, substantially reducing information loss and inter-head conflicts. The psuedo-code implementing our proposed MAFS is depicted in Algorithm~\ref{alg:mafs}. 

\begin{algorithm}[H]
\caption{MAFS: Multi-head Attention-based Feature Selection}
\label{alg:mafs}
\begin{algorithmic}[1]
\State \textbf{Input:} Feature matrix $X \in \mathbb{R}^{n \times d}$, target vector $y \in \mathbb{R}^n$, filter method set $F = \{f_1, f_2, \ldots, f_k\}$
\State \textbf{Output:} Selected feature subset indices

\State  \textbf{Module 1: Filter Weight Initialization}
\For{$j = 1$ to $k$} \Comment{Compute feature importance for filter method $j$}
    \State $w^{(j)} \leftarrow f_j(X, y)$ 
 
    \State $w_{\text{norm}}^{(j)} \leftarrow (w^{(j)} - \mu_j) / \sigma_j$   \Comment{Normalization processing}
\EndFor

\State \textbf{Module 2: Multi-head Attention Processing}
\State Initialize attention head parameters $\{W_h, b_h\}$ for $h = 1, \ldots, M$ where $M = |F|$
\For{epoch $= 1$ to max\_epochs}
    \For{$h = 1$ to $M$}
        
        \State $\boldsymbol{\alpha}^{(h)} \leftarrow W_L^{(h)T} \text{ReLU}\!\left( \cdots \text{ReLU}\!\left(W_1^{(h)T} \boldsymbol{w}^{(h)}_{\text{norm}} + b_1^{(h)}\right) \cdots \right) + b_L^{(h)}$
        
        \Comment{Attention weights}
        \State $X^{(h)} \leftarrow X \odot \alpha^{(h)}$ \Comment{Soft feature selection}
    \EndFor
    
\State $L_{\text{total}}^{(h)}= \begin{cases} 
-\sum_{i=1}^{n} \sum_{c=1}^{C} y_{ic} \log(\hat{y}_{ic}^{(h)}) + \lambda \cdot \sum_{k=1}^{d} \tau_k^{(h)} |\alpha_k^{(h)}|& \text{for classification tasks} \\
\frac{1}{n}\sum_{i=1}^{n} (y_i - \hat{y}_i^{(h)})^2 + \lambda \cdot \sum_{k=1}^{d} \tau_k^{(h)} |\alpha_k^{(h)}| & \text{for regression tasks}

\end{cases}$
\EndFor

\State \textbf{Module 3: Feature Reordering}
\State $S \leftarrow \emptyset$ 
\For{$h = 1$ to $M$}
    \State $S \leftarrow S \cup \text{TopK}(\boldsymbol{\alpha}^{(h)}, K)$ 
\EndFor

\State Train reordering model$(S, y)$ to obtain feature importance scores 
\State $\text{final\_features} \leftarrow$ Sort and select final feature subset
\State \textbf{return} final features
\end{algorithmic}
\end{algorithm}

\subsubsection{Module 1: Filter weight initialization module}
The filter weight initialization module aims to incorporate prior knowledge from multiple filter-based feature selection methods to provide stable and effective initial weights for the downstream multi-head attention module. It is designed to capture diverse feature–outcome relationships by employing multiple filter-based feature selection methods (e.g., SIS, BCor-SIS, and Kendall’s tau), each representing a distinct statistical perspective. The module extracts prior feature-importance information from a set of filter methods $F = \{f_1, f_2, ..., f_k\}$, where each filter method $f_j$ independently computes feature importance vectors $w^{(j)} = f_j(X, y) \in \mathbb{R}^d$. Here, $X \in \mathbb{R}^{n \times d}$ represents the feature matrix and $y \in \mathbb{R}^n$ represents the target vector. 

Each filter vector $w^{(j)}$ is then normalized: $w^{(j)}_{\text{norm}} =\frac{w^{(j)} - \mu_j}{\sigma_j}$, where $\mu_j$ and $\sigma_j$ denote the mean and standard deviation of weights from the $j$-th filter method, respectively. Normalization prevents small-range filter weights from producing degenerate attention inputs. Although normalization may slightly distort the original weight distribution\cite{qiu2005effects}, potentially weaken strong preference signals of certain filter methods, it ensures that each attention head in the multi-head attention module receives meaningful and comparable input signals while preserving the distinctive characteristics of each filter method. These normalized filter weights provide a structured starting point that replaces random initialisation, facilitating more stable and efficient optimisation in the following attention module.

\subsubsection{Module 2: The multi-head attention module}

The multi-head attention core builds upon the filter module to evaluate features from multiple perspectives in parallel. By assigning each head to focus on a distinct source of prior knowledge (e.g., weights from SIS, BCor-SIS, and Kendall’s tau), the model simultaneously focuses on linear, nonlinear, and rank-based dependencies. Unlike traditional filter methods that assess features individually, each head in the attention module considers the joint contributions of multiple features, thereby capturing higher-order interactions. By integrating outputs across heads, our multi-head design extracts both confirmatory and complementary signals of feature relevance, improving robustness and capturing multi-dimensional dependencies that single-head attention often misses.

Fig.~\ref{fig:attention} illustrates the architecture of the multi-head attention module. The number of attention heads equals the number of filter methods in Module 1 (i.e., $M = |F|$). Each head processes its assigned filter weights independently through a dedicated MLP, preventing interference between different prior knowledge sources. The transformed representations from all heads are fed into their respective feedforward networks, which aggregate sample-level information and generate feature-wise importance scores for each head. For the $h$-th attention head ($h \in \{1,2,\ldots,M\}$), the attention weights $\boldsymbol{\alpha}^{(h)}$ are computed through an $L$-layer MLP as:

\begin{equation}
\boldsymbol{\alpha}^{(h)} = W_L^{(h)T} \text{ReLU}\left(W_{L-1}^{(h)T} \text{ReLU}\left(\cdots \text{ReLU}\left(W_1^{(h)T} \mathbf{w}_{\text{norm}}^{(h)} + \mathbf{b}_1^{(h)}\right) \cdots + \mathbf{b}_{L-1}^{(h)}\right)\right) + \mathbf{b}_L^{(h)},
\end{equation}

\noindent where $\{W_l^{(h)}, \mathbf{b}_l^{(h)}\}_{l=1}^L$ represent the learnable parameters for the $h$-th head, with $L$ being a configurable hyperparameter. Let $\alpha_k^{(h)}$ denote the $k$-th element of $\boldsymbol{\alpha}^{(h)}$, representing the importance score of feature $k$ according to head $h$. 

The soft feature selection mechanism operates through element-wise multiplication:
\begin{equation}
\mathbf{X}^{(h)} = \mathbf{X} \odot \boldsymbol{\alpha}^{(h)},
\end{equation}
\noindent where $\odot$ denotes the Hadamard product. This approach allows continuous feature weighting rather than categorical selection, enabling the model to capture subtle importance relationships.  The resulting attention weight distributions $\boldsymbol{\alpha}^{(h)}$ provide a transparent and interpretable assessment of feature importance. By comparing weight distributions across heads, users can understand consistency and differences among various feature selection perspectives.

The overall loss for the $h$th head ($h \in \{1, 2, \ldots, M\}$) integrates the primary task loss with a feature selection regularization term:

\begin{equation}
L_{\text{total}}^{(h)} = L_{\text{pred}}^{(h)} + \lambda \sum_{k=1}^{d} \tau_k^{(h)} |\alpha_k^{(h)}|,
\end{equation}

\noindent where $L_{\text{pred}}^{(h)}$ denotes the prediction loss for head $h$  (cross-entropy for classification, mean squared error for regression), $d$ is the total number of features, $\lambda$ is the regularization strength controlling the trade-off between predictive performance and feature sparsity, and $\tau_k^{(h)}$ is a feature–head–specific adaptive weight that we designed following a similar idea in adaptive lasso. Specifically, to leverage prior knowledge from each filter method and encourage the non-relevant features to have zero effects , we defined $\tau_k^{(h)}$ based on the initial filter weights as:

\begin{equation}
\tau_k^{(h)} = \min\left(\frac{1}{(|w_{\text{norm},k}^{(h)}| + \epsilon)^\gamma}, \tau_{\max}\right),
\end{equation}

\noindent where $w_{\text{norm},k}^{(h)}$ is the $k$-th element of the normalized filter weight vector $\mathbf{w}_{\text{norm}}^{(h)}$ from Module 1, representing the normalized initial weight of the $k$-th feature from the filter method assigned to head $h$, $\epsilon$ is a small constant preventing numerical instability, $\gamma$ controls the adaptiveness of the penalty, and $\tau_{\max}$ caps the maximum penalty to prevent extreme values. This adaptive mechanism assigns smaller penalties to features with strong prior support (high $|w_{\text{norm},k}^{(h)}|$) and larger penalties to those with weak support, enabling each head to adjust feature importance according to its specific prior knowledge source while maintaining sparsity through $L_1$ regularization.

Note that although MAFS employs a multi-head architecture, the $M$ heads operate in parallel without fusion. Each head independently generates its own feature selection outcomes and predictions, allowing diverse feature subsets to be explored simultaneously. This design facilitates comparative analysis across different statistical perspectives and supports robustness assessment through an ensemble-like evaluation.

\subsubsection{Module 3: Feature Reordering Module}

The feature reordering module serves as the final optimization stage of MAFS. Its objective is to consolidate informative features identified across multiple attention heads while minimizing information loss. Traditional methods typically aggregate outputs from filter-based approaches through simple voting or averaging, but this strategy can miss features that are highly informative under only specific statistical criteria. To overcome this limitation, MAFS employs a candidate set construction and re-ranking strategy (Fig.~\ref{fig:rerank}), enabling the model to capture complementary signals from multiple perspectives without prematurely discarding potentially important features.

For each attention head $h \in \{1, \ldots, M\}$, we select the top $K$ features according to its learned attention weights. The candidate set $S$ is formed by merging selected features from all heads:

\begin{equation}
S = \bigcup_{h=1}^{M} \text{TopK}(\boldsymbol{\alpha}^{(h)}, K),
\end{equation}

\noindent where $\text{TopK}(\boldsymbol{\alpha}^{(h)}, K)$ returns the indices of the $K$ features with the highest attention weights in the corresponding head. The resulting candidate set satisfies $K \leq |S| \leq M \cdot K$, with the actual size depending on feature overlap across heads.

Features in $S$ are re-ranked using any algorithms capable of estimating feature importance from $(\mathbf{X}_S, \mathbf{y})$, where $\mathbf{X}_S$ denotes the data matrix restricted to features in $S$. Once ultra-high-dimensional features are filtered into the set $S$, most existing feature importance algorithms, including the wrapper-based methods, can be applied. Suitable methods can be model-based importance (e.g., Random Forest, ExtraTrees), perturbation-based methods (e.g., permutation importance, SHAP values). Note that, to avoid biased assessment, filter-based methods used in Module 1 should not be reused in this stage. In our implementation, we employed tree-based models and re-ranked the features based on Gini impurity:

\begin{equation}
\beta_j = \frac{1}{N_{\text{trees}}} \sum_{t=1}^{N_{\text{trees}}} \text{Gini}_j^{(t)},
\end{equation}

\noindent where $\text{Gini}_j^{(t)}$ denotes the decrease in Gini impurity attributed to feature $j$ in tree $t$. Features are ranked in descending order of $\{\beta_j\}$ to produce the final ranking $\mathcal{R}$, from which the top-$\ell$ features are selected according to application or validation criteria.

\subsection{Commonly used deep learning based feature selection methods}

We compared our method against four widely used deep learning-based feature selection approaches: GRACES\cite{chen2023graph}, EAR-FS\cite{xue2023external}, DeepLIFT \cite{shrikumar2017learning}, and CancelOut\cite{borisov2019cancelout}. These methods were selected for comparison as they represent diverse paradigms of deep learning–based feature selection, covering layer-based gating, gradient-based attribution, attention mechanisms, and graph-based learning.

CancelOut introduces a learnable gating layer based feature selection method\cite{borisov2019cancelout}. It integrates a customized feature selection layer into deep neural networks that automatically learns the importance weights of each input feature during model training. This layer filters input features through element-wise weighting, with its output defined as:

\begin{equation}
\text{CancelOut}(\mathbf{X}) = \mathbf{X} \odot \sigma(\mathbf{W}_{CO}),
\end{equation}

\noindent where $\mathbf{X}$ is the input matrix ($n \times p$), $\mathbf{W}_{CO}$ is the learnable weight vector, $\sigma(\cdot)$ is the sigmoid activation function, and $\odot$ denotes element-wise multiplication. During training, the CancelOut layer achieves feature selection by incorporating two regularization terms in the loss function to constrain the weight distribution:

\begin{equation}
\mathcal{L}_{\text{total}} = \mathcal{L}_{\text{pred}} - \lambda_1 \cdot \text{var}\left(\frac{\mathbf{W}_{CO}}{p}\right) + \lambda_2 \left\|\frac{\mathbf{W}_{CO}}{p}\right\|_1,
\end{equation}

\noindent where $\mathcal{L}_{\text{pred}}$ is the loss, $p$ is the number of features, and $\lambda_1$ and $\lambda_2$ control the weights of the variance term and sparsity term, respectively. After training, feature importance is estimated by applying the sigmoid activation to the learned weights $\mathbf{W}_{CO}$, and ranking features according to $\sigma(w_{CO,i})$. In all our analyses, we used default settings and treat $\lambda_1$ and $\lambda_2$ as the hyperparameters that require tuning.

DeepLIFT is a gradient-based attribution method that explains model predictions by decomposing the output into contributions from each input feature\cite{shrikumar2017learning}. The method computes feature importance scores as the difference between each neuron's activation and its reference activation, which are then propagated backwards through the network via modified backpropagation rules. Features are ranked based on the absolute magnitudes of their attribution scores. In our analyses, we used the default zero vector as the reference input.

EAR-FS is an attention-based feature selection method\cite{xue2023external}. It learns feature importance through an external attention mechanism embedded in a multilayer perceptron network. The attention module is a learnable vector $\mathbf{a} \in \mathbb{R}^{p}$ (where $p$ is the number of features), transformed by sigmoid activation to $\mathbf{a}^* = \sigma(\mathbf{a})$, which is element-wise multiplied with input features. During training, the network minimizes a loss function with a regularization term:

\begin{equation}
\mathcal{L}_{\text{total}} = \mathcal{L}_{\text{pred}} + \frac{\lambda}{\sum_i\left((a_i^* - 0.5)^2\right)},
\end{equation}

\noindent where the first term is the loss, $\lambda$ is the regularization coefficient, and the second term penalizes attention weights near 0.5 to encourage clear differentiation between important and unimportant features. After training, features are ranked by their attention values $a_i^*$. In our analyses, attention weights were initialized using the default setting and the hyperparameter $\lambda$ were tuned . 

GRACES is a graph-based feature selection method\cite{chen2023graph}. It learns feature importance through input gradients from a graph convolutional network (GCN) trained on a dynamically constructed sample similarity graph. During graph construction, sample pairs are connected only when their cosine similarity exceeds the $\alpha$-th percentile threshold. The GCN is trained by minimizing a task-specific loss, and feature importance is estimated through stochastic gradient estimation: the trained network undergoes $m=10$ independent dropout operations (dropout rate = 0.5), with Gaussian white noise added to the GraphSAGE layer weights in each realization. The averaged gradient matrix is computed as:
\begin{equation}
G = \frac{1}{m} \sum_{i=1}^{m} \frac{\partial \mathcal{L}}{\partial W_{\text{input}}^{(i)}},
\end{equation}
\noindent where $W_{\text{input}}^{(i)}$ denotes the input weights of the $i$-th dropout realization and $m=10$. The gradient-based feature score is obtained by computing each feature's gradient column: $g_j = \|G_{\cdot j}\|_2$. To enhance robustness, GRACES fuses gradient-based scores with ANOVA $F$-statistics $f_j$ through a weighted combination:
\begin{equation}
s_j = f_{\text{correct}} \cdot g_j + (1 - f_{\text{correct}}) \cdot f_j,
\end{equation}
\noindent where $f_{\text{correct}} \in [0,1]$ controls the relative contribution of gradient and statistical evidence. Features are iteratively selected by choosing $j^{*} = \arg\max_{j \notin S} s_j$ and retraining the network with the updated feature subset until the target number is reached. In our analysis, hyperparameters $\alpha$ and $f_{\text{correct}}$ were tuned.

\subsection{Simulation}
To evaluate the performance of MAFS, we conducted two simulation experiments: (1) The functional relationship analysis systematically evaluates how model performance varies across different feature-outcome relationships (linear, nonlinear, and interaction-based) under varying pre-specified feature selection ratios. (2) The dimensional scalability analysis examines model performance across varying feature dimensions.  Both simulations were designed to reflect realistic characteristics of biomedical data, including high dimensionality, heterogeneous feature types, and complex functional relationships.

We generated six feature–outcome combinations by crossing three feature types (continuous gene expression, categorical SNP genotypes, or combined) with two outcome types (continuous phenotype or binary disease status). This $3 \times 2$ design reflects common biomedical research scenarios: biomarker quantification with continuous features and outcomes, disease classification with continuous or combined features and categorical outcomes, and genotype–phenotype association with categorical features and outcomes.
\subsubsection{Simulated Data}

To reflect realistic settings, we simulated high-dimensional gene expression (continuous) and SNP (categorical) data using the OmicsSIMLA software\cite{chung2019multi}, a versatile simulator capable of generating realistic multi-omics datasets that preserve underlying data structures and inter-omic dependencies.

For SNP data, OmicsSIMLA requires external reference haplotypes that are generated using a sequence simulator. Here, we employed HAPGEN2\cite{su2011hapgen2} to simulate a reference panel based on the CEPH population from the 1000 Genomes Project. This reference panel captures realistic linkage disequilibrium patterns and allele frequencies distributions of European populations. Individual genotypes were simulated across all autosomal chromosomes, with diploid genotypes encoded as 0, 1, or 2 representing the number of minor alleles. Approximately 24 million SNPs were available across the autosomal genome. Standard quality control (QC) was applied and SNPs were excluded if they: (1) exhibited zero variance across all samples, or (2) showed high linkage disequilibrium ($r^2 > 0.2$) with nearby variants within a sliding window of 200 SNPs (step size of 5 SNPs). After QC, approximately 350,000 SNPs remained, corresponding to a filtering rate of 98.5\%.

For RNA-seq gene expression data,  OmicsSIMLA models read counts through negative binomial distributions with parameters learned from real transcriptomic data. Specifically, gene-wise mean expression and dispersion parameters were calibrated to match the breast invasive carcinoma (BRCA) cohort from The Cancer Genome Atlas (TCGA). This ensures that the simulated data capture authentic biological variability observed in real tissues. To model realistic regulatory relationships between genotype and gene expression, expression quantitative trait loci (eQTL) were incorporated. The eQTL catalog was derived from GTEx breast tissue\cite{gtex2020}. We retrieved significant cis-eQTLs meeting dual thresholds: permutation-based $p < 10^{-8}$ and beta-approximation permutation $p < 0.05$. After restricting to autosomal eQTLs and harmonizing alleles and coordinates with the CEPH-based SNP panel, the resulting gene-SNP mapping was supplied to OmicsSIMLA. This allows simulated gene expression to be explicitly modulated by the corresponding genotypes. Following standard RNA-seq preprocessing, we obtained normalized expression matrices and excluded genes with no variation across samples. The resulting dataset contained approximately 26,000 genes whose expression values exhibited approximately normal distributions after normalization. This is consistent with typical log-transformed RNA-seq data.

Feature-outcome relationships were modeled using seven different functional forms: linear, cosine, logarithmic, cubic, exponential, composite, and cosine-exponential interaction. The mean for each sample was defined in Equation ~\ref{eq:simulation}:
\begin{align}
\mu_i &= \beta_1 \sum_{a \in A} x_{i,a}
+ \beta_2 \sum_{b \in B} \cos(x_{i,b})
+ \beta_3 \sum_{c \in C} \log(|x_{i,c}| + 10^{-6}) + \beta_4 \sum_{d \in D} x_{i,d}^3 \nonumber\\
&\quad+ \beta_5 \sum_{e \in E} \exp(x_{i,e}) + \beta_6 \sum_{(g,k)\in (G,K)} \cos(x_{i,g_k}) \cdot \exp(x_{i,h_k})\nonumber\\
&\quad + \beta_7 \Big[\sum_{f \in F} \cos(x_{i,f})
+ \sum_{f \in F} \log(|x_{i,f}| + 10^{-6})
+ \sum_{f \in F} x_{i,f}^3 + \sum_{f \in F} \exp(x_{i,f}) \Big]
,
\label{eq:simulation}
\end{align}

\noindent 
where $\{A, \dots, H\}$ denotes disjoint feature index sets partitioning the causal features, with each set assigned to one functional form to avoid overlap. This design ensures independent evaluation of each relationship type. A small constant $10^{-6}$ was added for numerical stability in logarithmic transformations. For continuous or categorical feature scenarios, 40 causal features were selected from the available feature pool. These consisted of 5 features for each of the six non-interaction functional forms and 10 features for the interaction term. For combined scenarios involving both continuous and categorical features, 48 causal features were selected. This included 24 continuous features and 24 categorical features. Each feature type contributed 3 features to each of the six non-interaction forms and 6 features to the interaction term. We considered both continuous and categorical response variables. For continuous outcomes, we simulated as $Y_i \sim N(\mu_i,1)$.  For binary outcomes, to ensure a balanced case-control design, we first centered the mean (denoted as $\tilde{\mu}_i$) and generated the outcome through binomial sampling as $Y_i \sim \text{Binomial}(1, p_i)$, where $p_i = (1 + \exp(-\tilde{\mu}_i))^{-1}$. We considered two sample sizes in our simulations, reflecting small-scale exploratory studies ($n=500$) and moderate-scale investigations ($n=2000$). Effect sizes for different functional relationships were summarized in Supplementary Table S1.
\subsubsection{Simulation Scenarios}

To systematically assess the feature selection performance of MAFS, we conducted two complementary simulation experiments, each focusing on distinct evaluation aspects. In both simulations, coverage rate was used as the primary performance metric, defined as the proportion of truly causal features of each functional form correctly identified within the selected subset. To ensure robustness, each simulation setting was repeated 20 times. Data were split into training (80\%) and validation (20\%) subsets, where validation set is used for hyper-parameter tuning. The average coverage rate and computational resources that includes both feature selection and hyperparameter optimization were reported.

Simulation 1 was designed to evaluate the method’s ability to detect relationship-specific features. Multiple predefined thresholds (i.e, 0.5\%, 1\%, 1.5\% and 2\% of the total number of features) were applied to determine whether informative features of each functional form were prioritized early in the selection process or only recovered when larger subsets were considered. This analysis provided insights into the sensitivity and robustness of MAFS across diverse functional dependencies, as well as practical guidance on determining the number of top-ranked features for downstream analyses. 

Simulation 2 was designed to evaluate the scalability of MAFS under increasing feature dimensionalities. The total number of features was set to 25,000, 50,000, and 100,000, corresponding to typical transcriptomic/proteomic datasets, multi-omics integration scenarios, and ultra-high-dimensional genomic or epigenomic studies. Because the default outputs from OmicsSIMLA did not match the target dimensions, we applied dimension augmentation for continuous data and dimension truncation for categorical data. For continuous features, additional variables were generated by varying mean parameters and concatenating multiple feature sets when the desired dimension is larger than OmicsSIMLA's default. For categorical features, SNPs were sequentially selected from the quality-controlled set to match the desired dimension. For combined-type features, gene expression and SNP genotypes were concatenated at a 50\%/50\% ratio to achieve the target dimensions. Two feature selection strategies were employed for this set of simulation: a fixed-proportion strategy (i.e., 2\% of the total number of features), suitable for large-scale search spaces, and a fixed-count strategy (i.e., top 100 and 500 ranked features), which reflects the constraints of resource-limited validation studies. Together, these simulations evaluated the robustness and scalability of MAFS as the dimensionality of the feature space increased. 

Computational efficiency across different sample sizes ($n=500$ and $n=2,000$), feature dimensions (25K, 50K, and 100K), and outcome types (continuous and binary) was assessed by total running time (minutes). Notably, GRACES selects features iteratively rather than generating global importance scores, making it computationally infeasible to evaluate all features. Therefore, for GRACES, runtime was reported for selecting up to 2\% of the total input features. The computational cost of MAFS includes all three modules: (1) three filter-based methods for weight initialization, (2) a multi-head attention deep neural network and (3)  the reordering module implmeneting a tree algorithm.

\subsubsection{Parameters for Deep Learning based Feature Selection Method for Simulations}

\begin{figure}[H]
    \includegraphics[width=\textwidth]{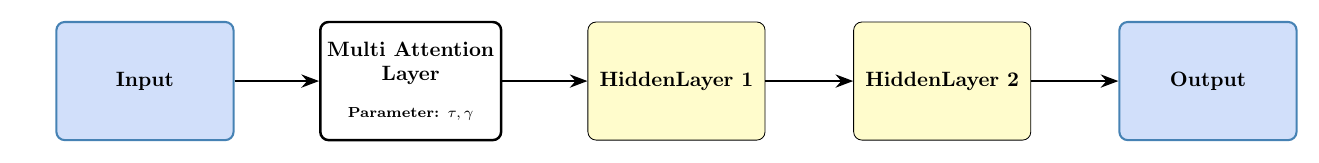}\\
    \includegraphics[width=\textwidth]{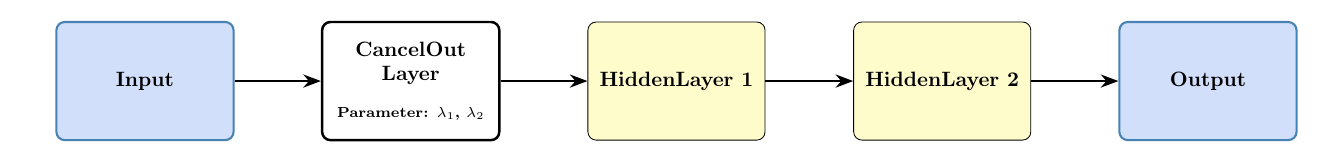}\\
    \includegraphics[width=\textwidth]{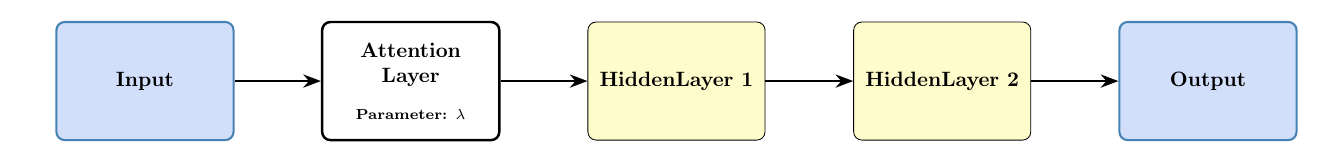}\\
    \includegraphics[width=\textwidth]{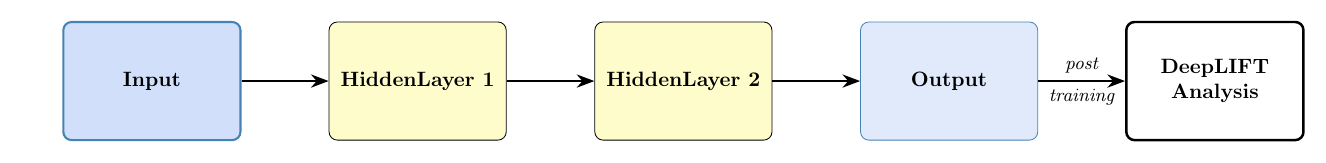}\\
    \includegraphics[width=\textwidth]{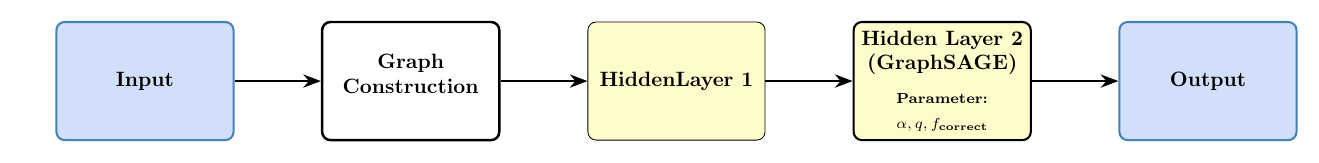}
    
    \caption{\textbf{Network architectures of baseline methods.} From top to bottom: MAFS, CancelOut, EAR-FS, DeepLIFT, and GRACES. All methods use comparable capacities with method-specific components highlighted.}
    \label{fig:baseline_architectures}
\end{figure}

Fig.~\ref{fig:baseline_architectures} illustrates the network architectures of all methods. To ensure fair comparison, all embedded feature selection methods (CancelOut, EAR-FS, GRACES, and MAFS) were implemented using a consistent network architecture. Each model employed a feedforward neural network with two hidden layers. The width of each layer was determined by dividing the input dimension by a scaling factor. Each layer consisted of a linear transformation, batch normalization, ReLU activation, and dropout. For DeepLIFT, which is a post-hoc attribution method, we first trained a feedforward network using the same architecture and then applied DeepLIFT to compute feature importance scores.

All models were trained using the Adam optimizer with a batch size of 32 and a maximum of 100 training epochs. Early stopping was applied when validation performance did not improve for 10 consecutive epochs. Key hyperparameters, including method-specific tuning parameters, learning rate, dropout rate, batch size, and scaling factor, were optimized using Optuna with 100 search iterations for each method and each data configuration\cite{akiba2019optuna}. Detailed hyperparameter search spaces and optimal values are provided in Supplementary Table S2 .

\subsection{Real Data Applications}

To assess the practical utility of MAFS, we analyzed two representative categories of real-world datasets. The first comprised six publicly available gene expression cancer datasets with binary outcomes (e.g., glioma vs. control), which serve as established benchmarks for classification-based feature selection\cite{chen2023graph}. The second category drew on the Alzheimer’s Disease Neuroimaging Initiative (ADNI) multi-omics dataset with continuous response variables, enabling assessment of feature selection performance in regression settings \cite{adni_website}.

\subsubsection{Cancer Gene Expression Datasets}
The six cancer gene expression datasets are widely used as benchmark datasets for feature selection. They include both hematological malignancies and solid tumors, spanning a broad range of sample sizes (62–187) and dimensionalities (2,000–22,283). Feature types include both continuous measurements (normalized expression values as real numbers) and discrete variables (ordinal-encoded categories). All datasets represent binary classification tasks. Detailed characteristics are provided in Supplementary Table S4. Each dataset was obtained in its preprocessed form from the corresponding original publication or GEO repository and contains no missing values. We used the data as provided, without performing additional preprocessing.

\subsubsection{Alzheimer's Disease Neuroimaging Initiative Dataset}

ADNI is a multicenter longitudinal study established to identify and validate clinical, imaging, genetic, and biochemical biomarkers for the early detection and monitoring of Alzheimer's disease. In our study, we focused on selecting predictive gene expression features for imaging-derived volumetric phenotypes, including the accumbens, amygdala, brainstem, caudate, fourth ventricle, hippocampus, pallidum, putamen and thalamus.

Imaging-derived volumetric phenotypes and gene expression features were used from ADNI website\cite{adni_website}. To ensure independent observations, we restricted analyses to baseline visits when multiple records were available for a participant. We retained only samples with RNA Integrity Number (RIN) $\geq 6$ to ensure adequate RNA quality for reliable transcriptomic measurements\cite{fleige2006rna}. Participants with missing values in either gene expression or phenotypic data were excluded from analysis. The final analysis dataset consisted of 449 participants with 49,386 gene expression features.

\subsubsection{Parameters for Deep Learning based Feature Selection Method and Evaluation Metrics for Real Data Applications}

To evaluate the feature selection capability of MAFS on real-world datasets, we employed the same set of comparison methods as in the simulation data. All deep learning–based feature selection methods adopted the same network architectures (Fig.~\ref{fig:baseline_architectures}), training settings, and hyperparameter search ranges as in the simulation studies, with model-specific hyperparameters optimised via Optuna (100 iterations). Detailed descriptions of the search spaces are provided in Supplementary Table S2.

Because the true relevant features are unknown in real-world datasets, prediction accuracy was used as a proxy to assess feature selection performance. In the cancer datasets, following standard benchmarking practice\cite{chen2023graph}, the top 20 ranked features were retained for downstream classification analysis. Three classifiers were employed: support vector machine (SVM), k-nearest neighbors (KNN), and a multilayer perceptron (MLP) with two hidden layers (64 and 32 units) and ReLU activation. Classification performance was evaluated using the area under the receiver operating characteristic curve (AUROC). Considering the limited sample sizes of the cancer gene expression datasets, we employed a 5-fold cross-validation framework to maximize training data utilization. Each dataset was split with 20\% reserved as the test set and the remaining 80\% used for feature selection and model training.

For each imaging-derived outcome, the top 50 ranked features were retained for regression analysis. This threshold balances model complexity and predictive performance while avoiding overfitting in the moderate-sample-size setting ($n=449$). Three regressors were employed: support vector regression (SVR), KNN, and an MLP with the same two-hidden-layer architecture (64 and 32 neurons, ReLU activation). Predictive performance was evaluated using the Pearson correlation coefficient between predicted and observed values. Given the sufficient sample size, a fixed 60:20:20 train–validation–test split was adopted, and performance on the test set was reported.

Hyperparameters for all downstream prediction models were optimised via grid search (Supplementary Table S3). To ensure the robustness of the findings, all experiments were repeated 20 times with identical data partitions and random seeds across methods.

\section*{Data availability}

The six cancer gene expression datasets (Colon, Leukemia, ALLAML, GLI\_85, Prostate\_GE, and SMK\_CAN\_187) used in this study are publicly available benchmark datasets that can be accessed from \url{https://jundongl.github.io/scikit-feature/datasets.html}.  Imaging-derived volumetric phenotypes and gene expression data from the Alzheimer's Disease Neuroimaging Initiative (ADNI) were obtained from the ADNI database (\url{https://adni.loni.usc.edu/}). Access to ADNI data requires registration and approval through the ADNI Data Archive. Simulation data generated for this study and source data underlying the figures are available from the corresponding author upon reasonable request.

\section*{Code availability}

All code used to implement MAFS and reproduce the results in this study is publicly available at \url{https://github.com/xsun768/mafs_package}. The implementation is written in Python and includes scripts for  model training and performance evaluation. Dependencies and installation instructions are provided in the repository.

\section*{Acknowledgments}
The author(s) wish to acknowledge use of the eResearch Infrastructure Platform hosted by the Crown company, Research and Education Advanced Network New Zealand (REANNZ) Ltd., and funded by the Ministry of Business, Innovation \& Employment. URL: https://www.reannz.co.nz

\section*{Author Contributions}
X.S. and Q.M. contributed equally to this work. X.S. conceived the idea, developed the algorithm, performed simulations, and analyzed real data. Q.M. assisted with simulation data generation and real data analyzed. X.S. and Y.W. wrote the manuscript. Y.W. supervised the work.

\subsection*{Corresponding Author}
Correspondence and requests for materials should be addressed to Yalu Wen.

\section*{Competing interests}
The authors declare no competing interests.

\end{document}


\setlength{\parindent}{2em}

\newpage
\appendix
\begin{center}
{\Large \textbf{Supplementary File for ``MAFS: Multi-head Attention Feature Selection for High-Dimensional Data via Deep Fusion of Filter Methods''}}

\textbf{Xiaoyan Sun}\textsuperscript{1,†}, \textbf{Qingyu Meng}\textsuperscript{1,†}, \textbf{Yalu Wen}\textsuperscript{1,*}

\vspace{1cm}

\textsuperscript{1}Department of Statistics, University of Auckland, Auckland, New Zealand

\textsuperscript{†}These authors contributed equally to this work.

\textsuperscript{*}To whom correspondence should be addressed: y.wen@auckland.ac.nz
\end{center}

\setcounter{figure}{0} 
\renewcommand{\thefigure}{S\arabic{figure}}

\setcounter{table}{0}
\renewcommand{\thetable}{S\arabic{table}}

\section{Supplementary Tables}

\begin{table}[htbp]
\centering
\small
\begin{tabular}{lcccccccc}
\toprule
\multicolumn{1}{c}{\textbf{Data Type Combination}} &
\multicolumn{7}{c}{\textbf{Relationship}} \\
\cmidrule(lr){2-8}
& \textbf{Linear} & \textbf{Cosine} & \textbf{Log} & \textbf{Cubic} & \textbf{Exp} & \textbf{Combined} & \textbf{Interaction} \\
\toprule
\multicolumn{8}{l}{\textbf{n = 500}} \\
\midrule
continuous X, categorical Y & 1.5 & 3.0 & 2.0 & 0.5 & 1.0 & 0.4 & 1.0 \\
continuous X, continuous Y   & 1.5 & 4.0 & 3.0 & 0.7 & 1.2 & 0.4 & 1.2 \\
categorical X, categorical Y & 1.5 & 3.0 & 0.15 & 1.5 & 1.0 & 0.4 & 1.2 \\
categorical X, continuous Y   & 1.5 & 3.0 & 0.15 & 1.5 & 1.2 & 0.4 & 1.0 \\
\addlinespace
\textit{Combined X, categorical Y} \\
\quad -continuous component   & 1.5 & 4.0 & 2.0 & 0.5 & 0.8 & 0.3 & 1.0  \\
\quad -categorical component & 1.5 & 4.0 & 0.4 & 1.5 & 0.8 & 0.3 & 1.0 \\
\addlinespace
\textit{Combined X, continuous Y} \\
\quad -continuous component   & 3.0 & 4.0 & 3.0 & 0.5 & 0.8 & 0.4 & 1.5 \\
\quad -categorical component & 3.0 & 4.0 & 0.3 & 0.8 & 1.2 & 0.5 & 1.5 \\
\midrule
\addlinespace[1em]
\multicolumn{8}{l}{\textbf{n = 2,000}} \\
\midrule
continuous X, categorical Y & 0.4 & 1.5 & 0.7 & 0.15 & 0.2 & 0.08 & 0.25 \\
continuous X, continuous Y   & 0.3 & 1.0 & 1.0 & 0.12 & 0.15 & 0.05 & 0.25 \\
categorical X, categorical Y & 0.4 & 1.0 & 0.1 & 0.6 & 0.6 & 0.15 & 0.3 \\
categorical X, continuous Y   & 1.0 & 1.0 & 0.05 & 0.5 & 0.3 & 0.1 & 0.15 \\
\addlinespace
\textit{Combined X, categorical Y} \\
\quad -continuous component   & 0.2 & 1.0 & 0.5 & 0.1 & 0.08 & 0.04 & 0.15 \\
\quad -categorical component & 0.3 & 1.0 & 0.05 & 0.4 & 0.15 & 0.04 & 0.2 \\
\addlinespace
\textit{Combined X, continuous Y} \\
\quad -continuous component   & 0.2 & 1.0 & 0.5 & 0.05 & 0.08 & 0.04 & 0.15 \\
\quad -categorical component & 0.3 & 1.0 & 0.05 & 0.3 & 0.15 & 0.02 & 0.2 \\
\bottomrule
\end{tabular}
\caption{\textbf{Effect sizes for simulation studies.} Effect sizes ($\beta$) for seven feature-outcome relationship types across data type combinations and sample sizes in simulation studies.}
\label{tab:coefficients}
\end{table}


\newpage



\begin{table}[htbp]
\centering
\small
\begin{tabularx}{\textwidth}{lXXX}
\toprule
\multirow{2}{*}{\textbf{Parameter}} &
\multicolumn{3}{c}{\textbf{Datasets}} \\
\cmidrule(lr){2-4}
& \textbf{Simulation} & \textbf{Cancer} & \textbf{ADNI} \\
\midrule
\multicolumn{4}{l}{\textbf{Common Parameters}} \\
\midrule
Learning rate 
& $[10^{-6}, 10^{-4}]$ 
& $[10^{-6}, 10^{-2}]$
& $[10^{-6}, 10^{-3}]$ \\
Batch size 
& $32$
& $\{4, 8, 16, 32\}$
& $\{8, 16, 32\}$ \\
Hidden dimension 
& $200$
& $[64, 512]$
& $[64, 512]$ \\
Dropout rate
& $0.4$
& $[0.2, 0.6]$
& $[0.3, 0.6]$ \\
Weight decay
& $[10^{-6}, 10^{-4}]$
& $[10^{-6}, 10^{-2}]$
& $[10^{-6}, 10^{-3}]$ \\
\midrule
\addlinespace[1em]
\multicolumn{4}{l}{\textbf{Specific Parameters}} \\
\midrule
\textbf{MAFS} & & & \\
\quad $\tau$ & $[10^{-6}, 10^{-4}]$ & $[10^{-6}, 10^{-2}]$ & $[10^{-6}, 10^{-3}]$ \\
\quad $\gamma$ & $[0.1, 0.5]$ & $[0.1, 0.5]$ & $[0.1, 1.0]$ \\
\addlinespace
\textbf{CancelOut} & & & \\
\quad $\lambda_{1}$ & $[10^{-5}, 10^{-1}]$ & $[10^{-4}, 10^{-1}]$ & $[10^{-5}, 10^{-1}]$ \\
\quad $\lambda_{2}$ & $[10^{-5}, 10^{-1}]$ & $[10^{-4}, 10^{-1}]$ & $[10^{-5}, 10^{-1}]$ \\
\addlinespace
\textbf{EAR-FS} & & & \\
\quad $\lambda$ & $[10^{-6}, 10^{-4}]$ & $[10^{-6}, 10^{-2}]$ & $[10^{-6}, 10^{-3}]$ \\
\addlinespace
\textbf{GRACES} & & & \\
\quad $\alpha$ & $[0.8, 0.99]$ & $[0.8, 0.99]$ & $[0.8, 0.99]$ \\
\quad $f_{\text{correct}}$ & $\{0, 0.1, 0.5, 0.9\}$ & $\{0, 0.1, 0.5, 0.9\}$ & $\{0, 0.1, 0.5, 0.9\}$ \\
\bottomrule
\end{tabularx}
\caption{\textbf{Hyperparameter ranges for feature selection methods.}  Hyperparameter ranges for all methods across experimental datasets. Common parameters apply to all deep learning-based methods. Method-specific parameters: MAFS---$\tau$ controls sparsity, $\gamma$ modulates penalty strength; CancelOut---$\lambda_{1}$ and $\lambda_{2}$ regulate L1 and variance penalties; EAR-FS---$\lambda$ controls attention regularization; GRACES---$\alpha$ controls graph consistency, $f_{\text{correct}}$ adjusts correction strength. DeepLIFT has no method-specific hyperparameters.}
\label{tab:hyperparameter}
\end{table}

\newpage

The same search grids were applied to the six cancer gene expression datasets and the ADNI neuroimaging dataset.

\begin{table}[htbp]
\centering
\begin{tabular}{lll}
\toprule
\textbf{Classifier} & \textbf{Hyperparameter} & \textbf{Search Range} \\
\midrule
\multirow{3}{*}{SVM} 
 & $C$ & \{0.1, 1, 10, 100\} \\
 & $\gamma$ & \{\texttt{'scale'}, \texttt{'auto'}, 0.01, 0.1\} \\
 & Kernel & \{\texttt{'rbf'}, \texttt{'linear'}\} \\
\midrule
\multirow{3}{*}{KNN} 
 & Number of neighbours ($n_{\text{neighbors}}$) & \{3, 5, 7, 9, 11, 15\} \\
 & Weights & \{\texttt{'uniform'}, \texttt{'distance'}\} \\
 & Distance metric & \{\texttt{'euclidean'}, \texttt{'manhattan'}, \texttt{'minkowski'}\} \\
\midrule
\multirow{3}{*}{MLP} 
 & Dropout probability & \{0.1, 0.3, 0.5\} \\
 & Learning rate & \{0.01, 0.001\} \\
 & $L_2$ regularisation ($\lambda$) & \{0.0001, 0.001, 0.01\} \\
\bottomrule
\end{tabular}
\caption{\textbf{Classifier hyperparameter search grids.} Hyperparameter search grids used for classifier evaluation following feature selection. These grids were applied consistently across the six cancer gene expression datasets and the ADNI neuroimaging dataset to ensure fair performance comparison. $C$ and $\gamma$ control margin width and kernel smoothness in SVM; $k$-nearest neighbour count ($n_{\text{neighbors}}$) and distance metric define KNN structure; and the MLP grid tunes dropout regularisation, learning rate, and $L_2$ penalty strength.}
\label{tab:hyperparams_realdata}
\end{table}


\begin{table}[h]
\centering
\begin{tabular}{lcccl}
\hline
\textbf{Dataset} & \textbf{Samples} & \textbf{Features} & \textbf{Feature Type} & \textbf{Target Type} \\
\hline
Colon\cite{alon1999broad} & 62 & 2,000 & Categorical & tumor(40) vs normal tissue(22) \\
Leukemia \cite{golub1999molecular}  & 72 & 7,129 & Categorical & acute lymphoblastic leukemia(47) vs AML(25) \\
ALLAML \cite{golub1999molecular} & 72 & 7,129 & continuous & acute lymphoblastic leukemia(47) vs AML(25) \\
GLI\_85 \cite{freije2004geneexpression} & 85 & 22,283 & continuous & Grade III(26) vs Grade IV gliomas(59) \\
Prostate\_GE \cite{singh2002geneexpression} & 102 & 5,966 & continuous & cancer(50) vs healthy controls(52) \\
SMK\_CAN\_187 \cite{spira2007airway} & 187 & 19,993 & continuous & smokers with(90) vs without lung cancer(97) \\
\hline
\end{tabular}
\caption{\textbf{Cancer gene expression dataset characteristics.} Characteristics of six cancer gene expression datasets used for real-world validation. Feature types include both continuous (continuous gene expression values) and categorical (discretized expression levels).}
\label{tab:cancer}
\end{table}


\section{Supplementary Figures}

\begin{figure}[htbp] 
    \centering
    \includegraphics[width=0.95\textwidth]{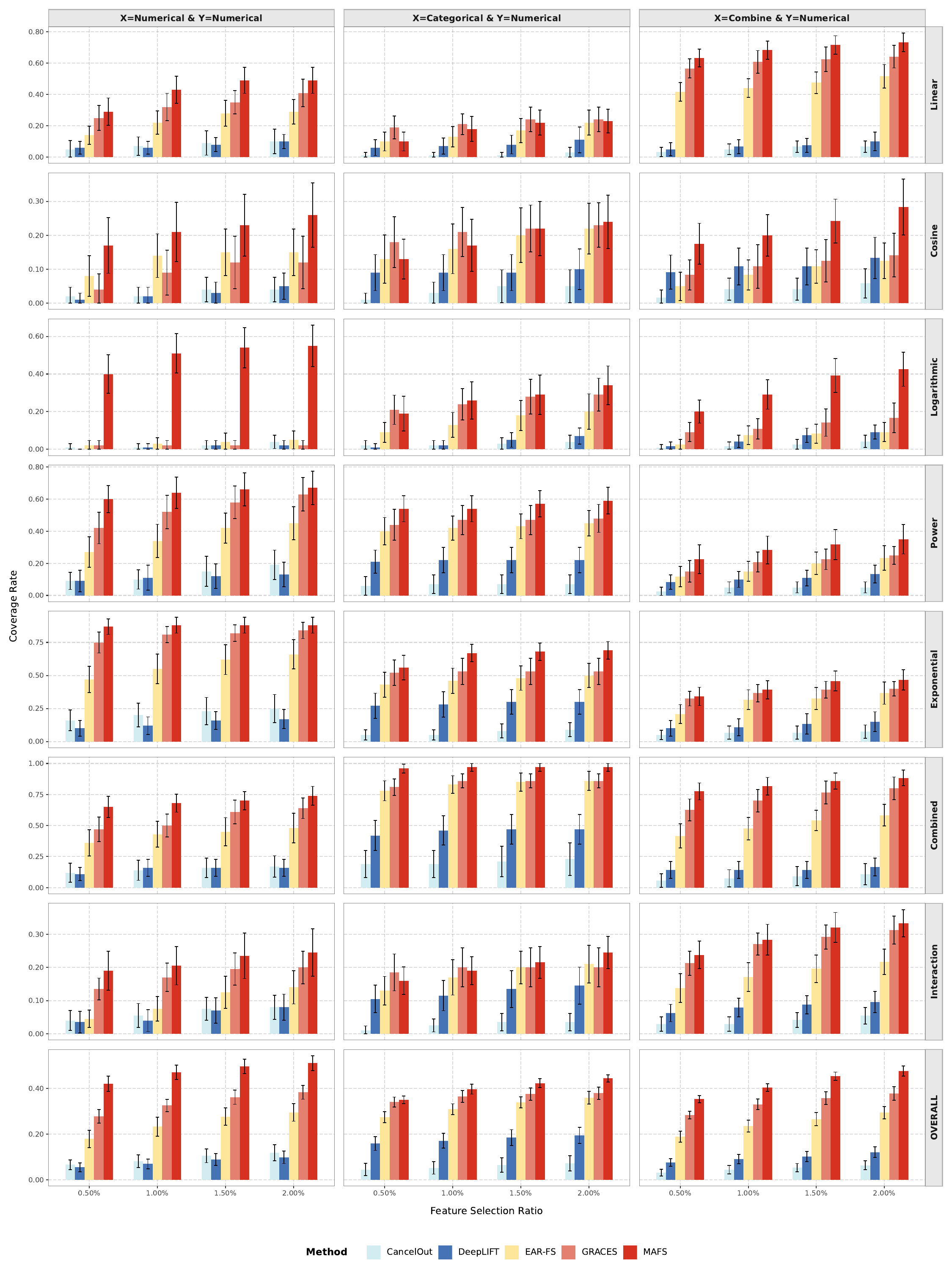}
    \caption{\textbf{Supplementary result of relationship-specific feature selection for continuous outcomes.} Coverage rates under seven different feature–response relationships at selection ratios 0.5\%, 1\%, 1.5\% and 2\% in the high-dimensional setting ($n=500$, $p=25{,}000$) for continuous outcomes.}
    \label{fig:relationship_analysis_500_25k}
\end{figure}

\begin{figure}[htbp] 
    \centering
    \includegraphics[width=0.95\textwidth]{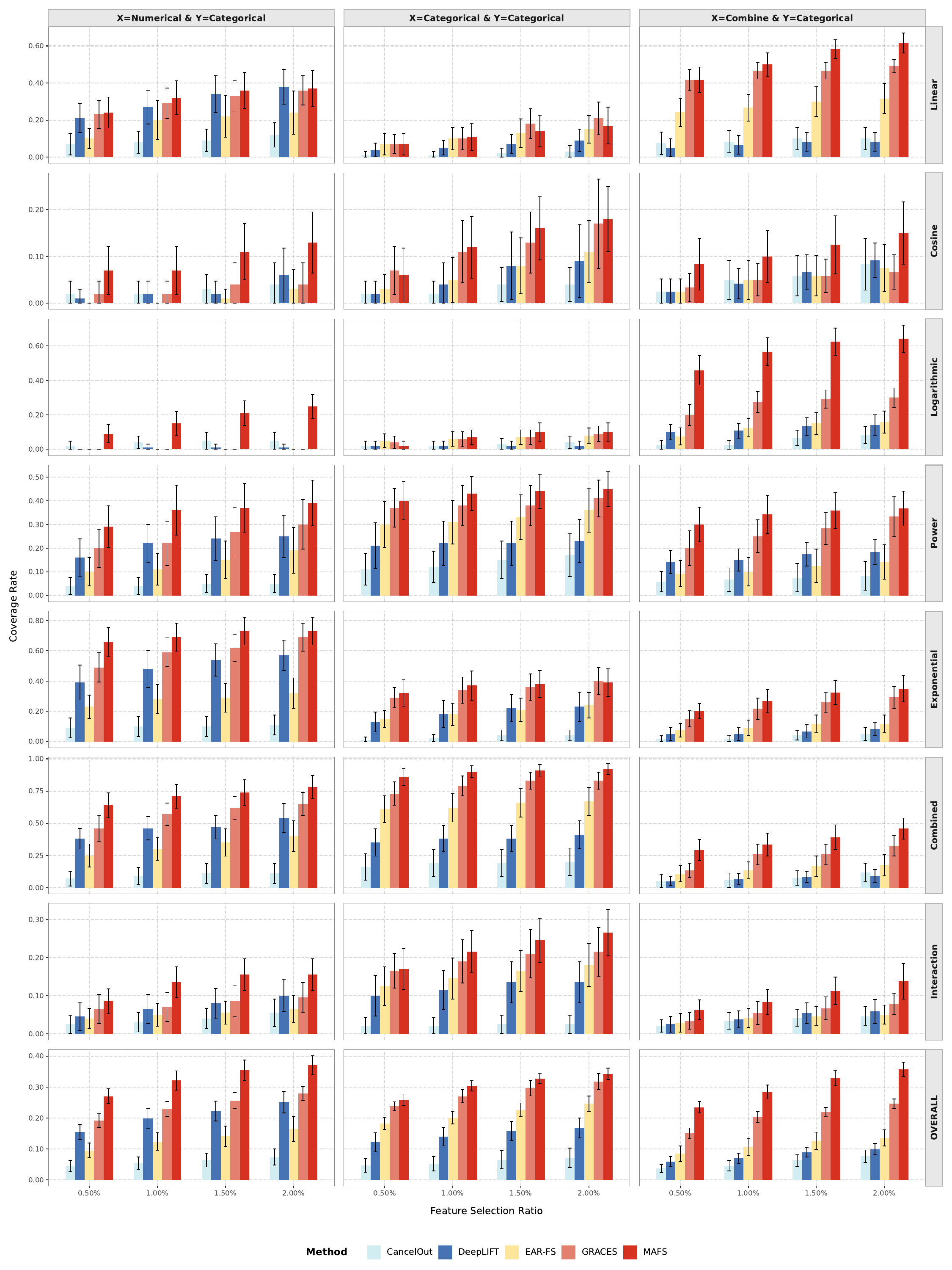}
    \caption{\textbf{Supplementary result of relationship-specific feature selection for categorical outcomes.} Coverage rates under seven different feature–response relationships at selection ratios 0.5\%, 1\%, 1.5\% and 2\% in the high-dimensional setting ($n=500$, $p=25{,}000$) for categorical outcomes.}
    \label{fig:relationship_analysis_500_25k_cate}
\end{figure}

\begin{figure}[htbp] 
    \centering
    \includegraphics[width=0.95\textwidth]{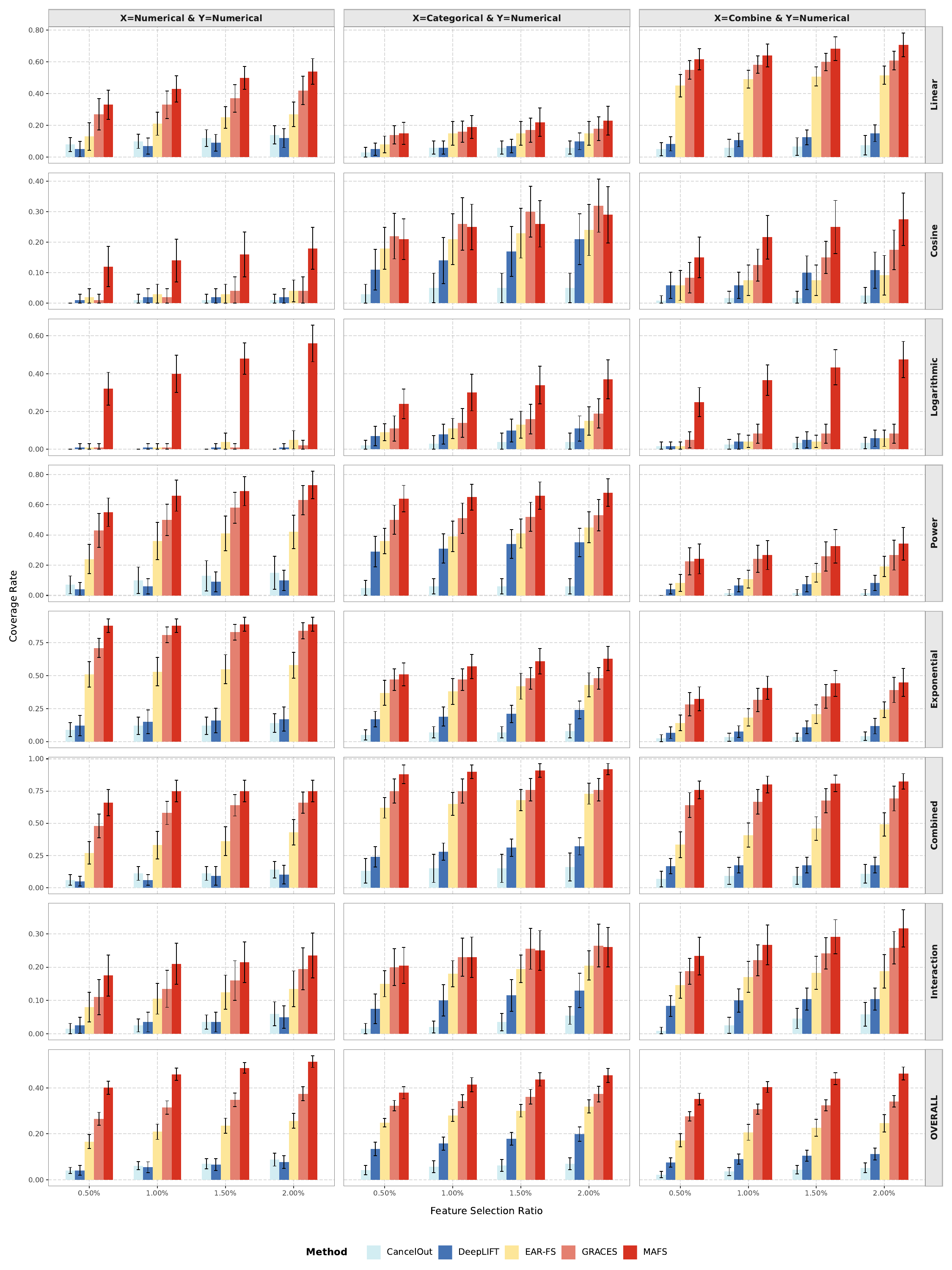}
    \caption{\textbf{Supplementary result of relationship-specific feature selection for continuous outcomes.} Coverage rates under seven different feature–response relationships at selection ratios 0.5\%, 1\%, 1.5\% and 2\% in the high-dimensional setting ($n=500$, $p=50{,}000$) for continuous outcomes.}
    \label{fig:relationship_analysis_500_50k}
\end{figure}

\begin{figure}[htbp] 
    \centering
    \includegraphics[width=0.95\textwidth]{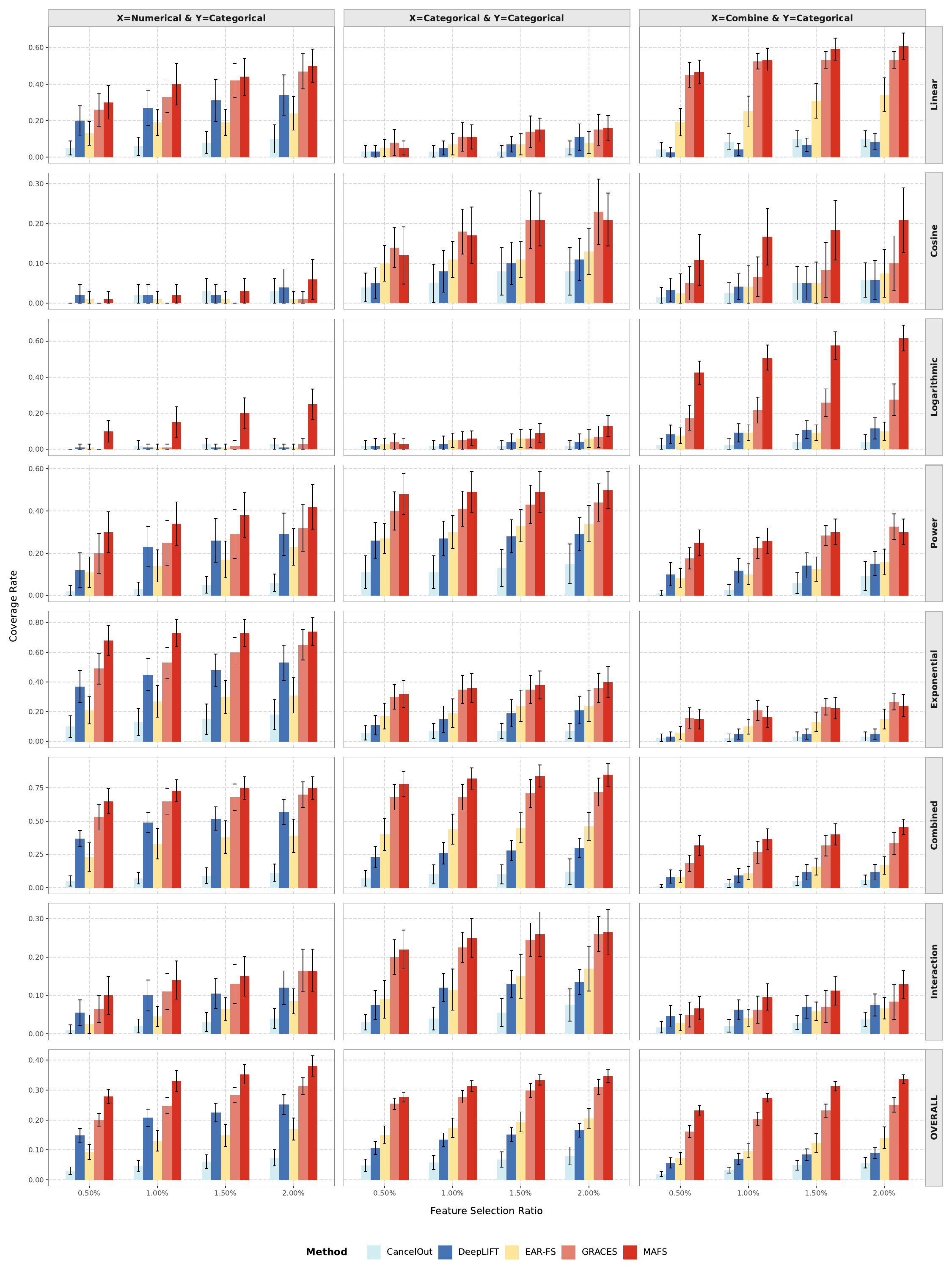}
    \caption{\textbf{Supplementary result of relationship-specific feature selection for categorical outcomes.} Coverage rates under seven different feature–response relationships at selection ratios 0.5\%, 1\%, 1.5\% and 2\% in the high-dimensional setting ($n=500$, $p=50{,}000$) for categorical outcomes.}
    \label{fig:relationship_analysis_500_50k_cate}
\end{figure}

\begin{figure}[htbp] 
    \centering
    \includegraphics[width=0.95\textwidth]{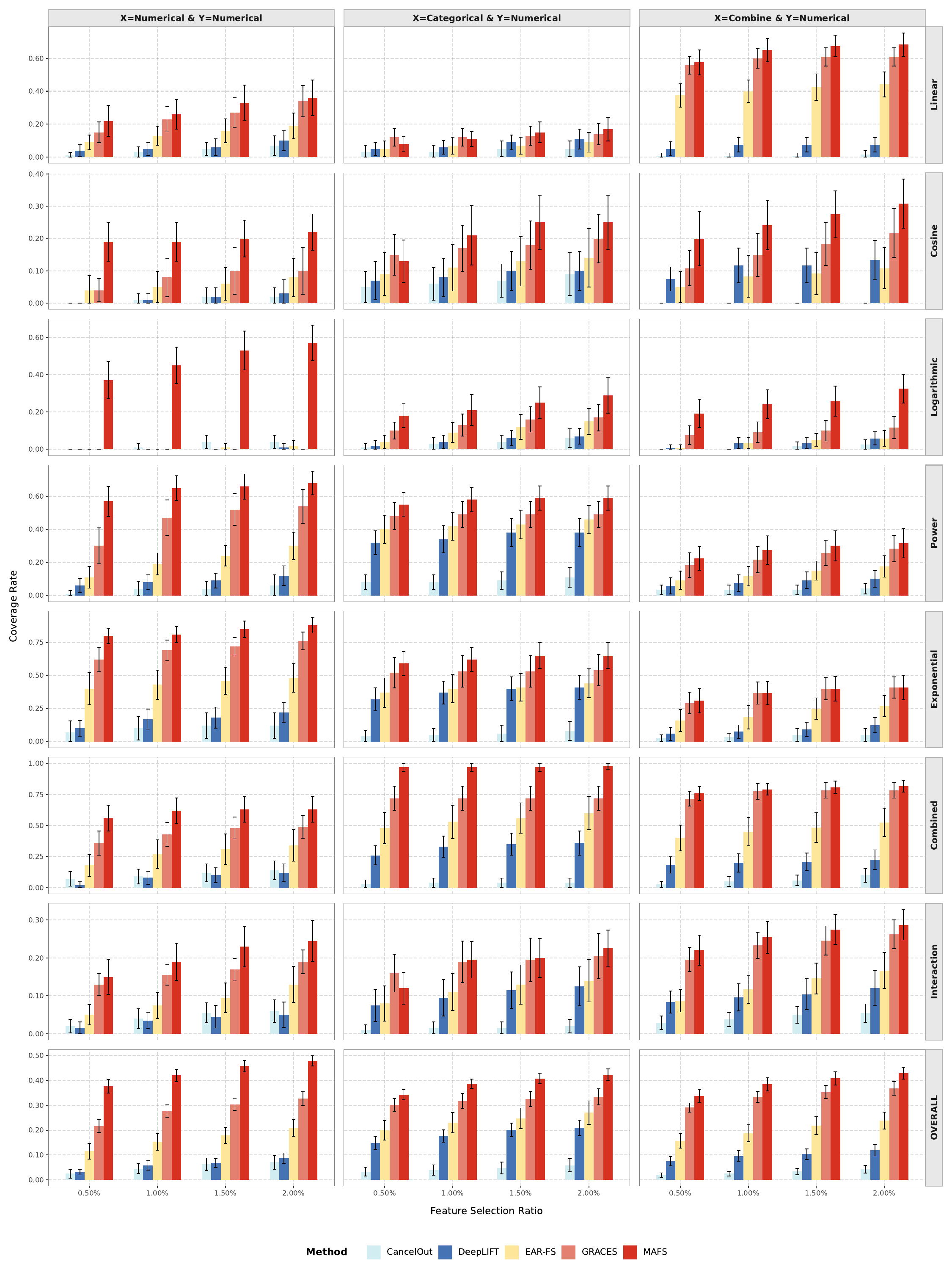}
        \caption{\textbf{Supplementary result of relationship-specific feature selection for continuous outcomes.} Coverage rates under seven different feature–response relationships at selection ratios 0.5\%, 1\%, 1.5\% and 2\% in the high-dimensional setting ($n=500$, $p=100{,}000$) for continuous outcomes.}
    \label{fig:relationship_analysis_500}
\end{figure}

\begin{figure}[htbp] 
    \centering
    \includegraphics[width=0.95\textwidth]{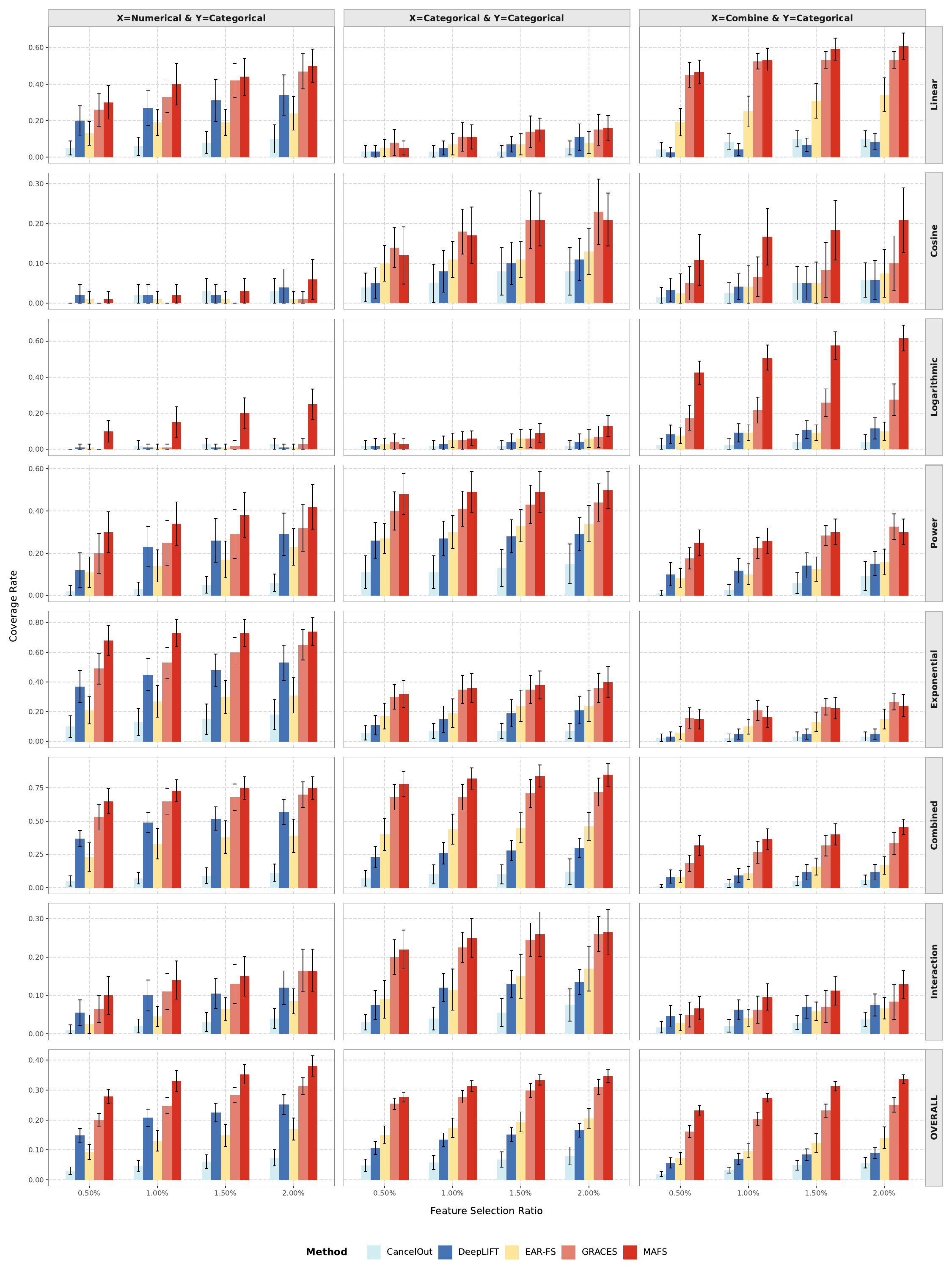}
    \caption{\textbf{Supplementary result of relationship-specific feature selection for categorical outcomes.} Coverage rates under seven different feature–response relationships at selection ratios 0.5\%, 1\%, 1.5\% and 2\% in the high-dimensional setting ($n=500$, $p=100{,}000$) for categorical outcomes.}
    \label{fig:relationship_analysis_500_cate}
\end{figure}


\begin{figure}[htbp] 
    \centering
    \includegraphics[width=0.95\textwidth]{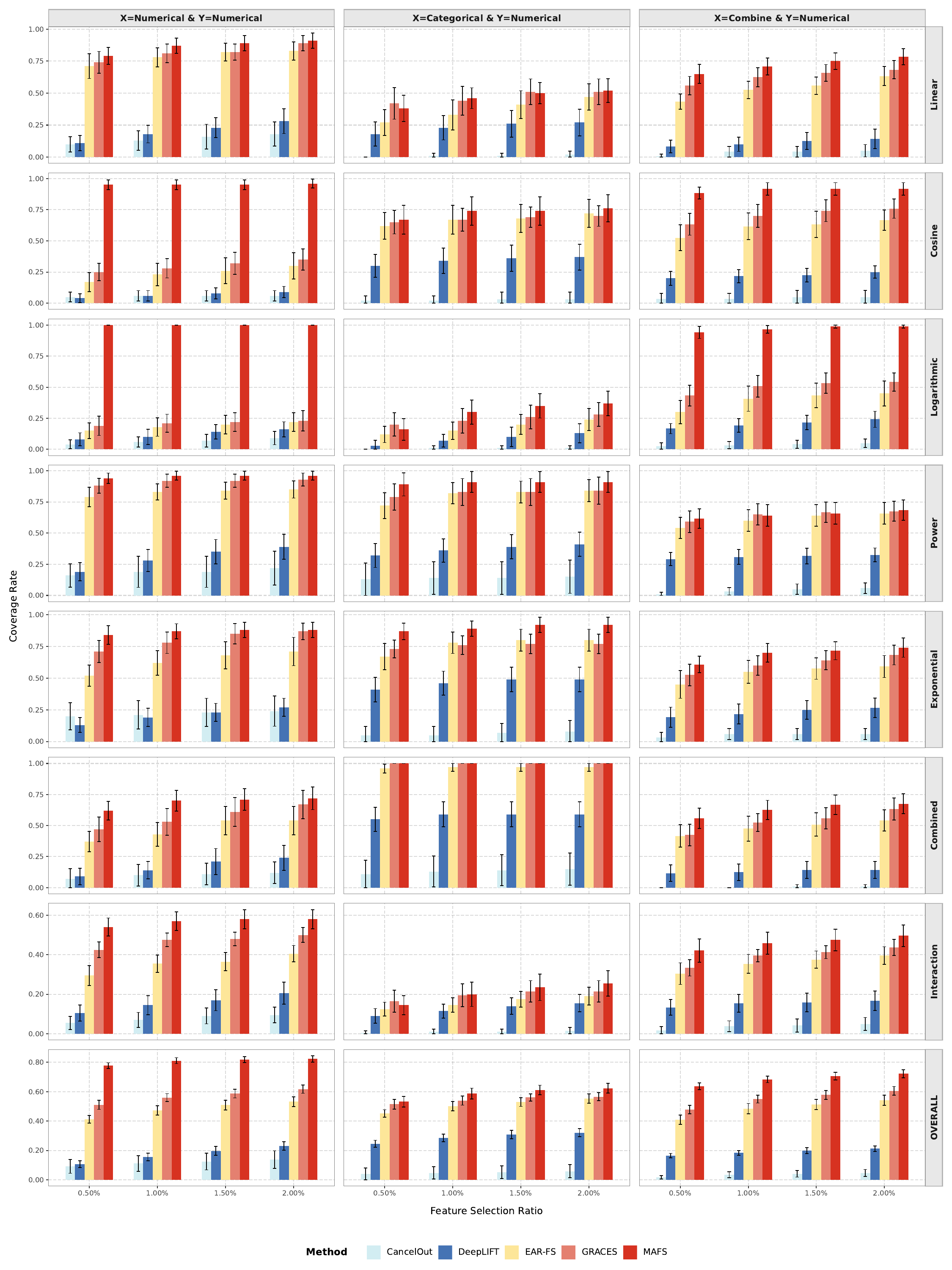}
    \caption{\textbf{Supplementary result of relationship-specific feature selection for continuous outcomes.} Coverage rates under seven different feature–response relationships at selection ratios 0.5\%, 1\%, 1.5\% and 2\% in the high-dimensional setting ($n=2{,}000$, $p=25{,}000$) for continuous outcomes.}
    \label{fig:relationship_analysis_2k_25k}
\end{figure}

\begin{figure}[htbp] 
    \centering
    \includegraphics[width=0.95\textwidth]{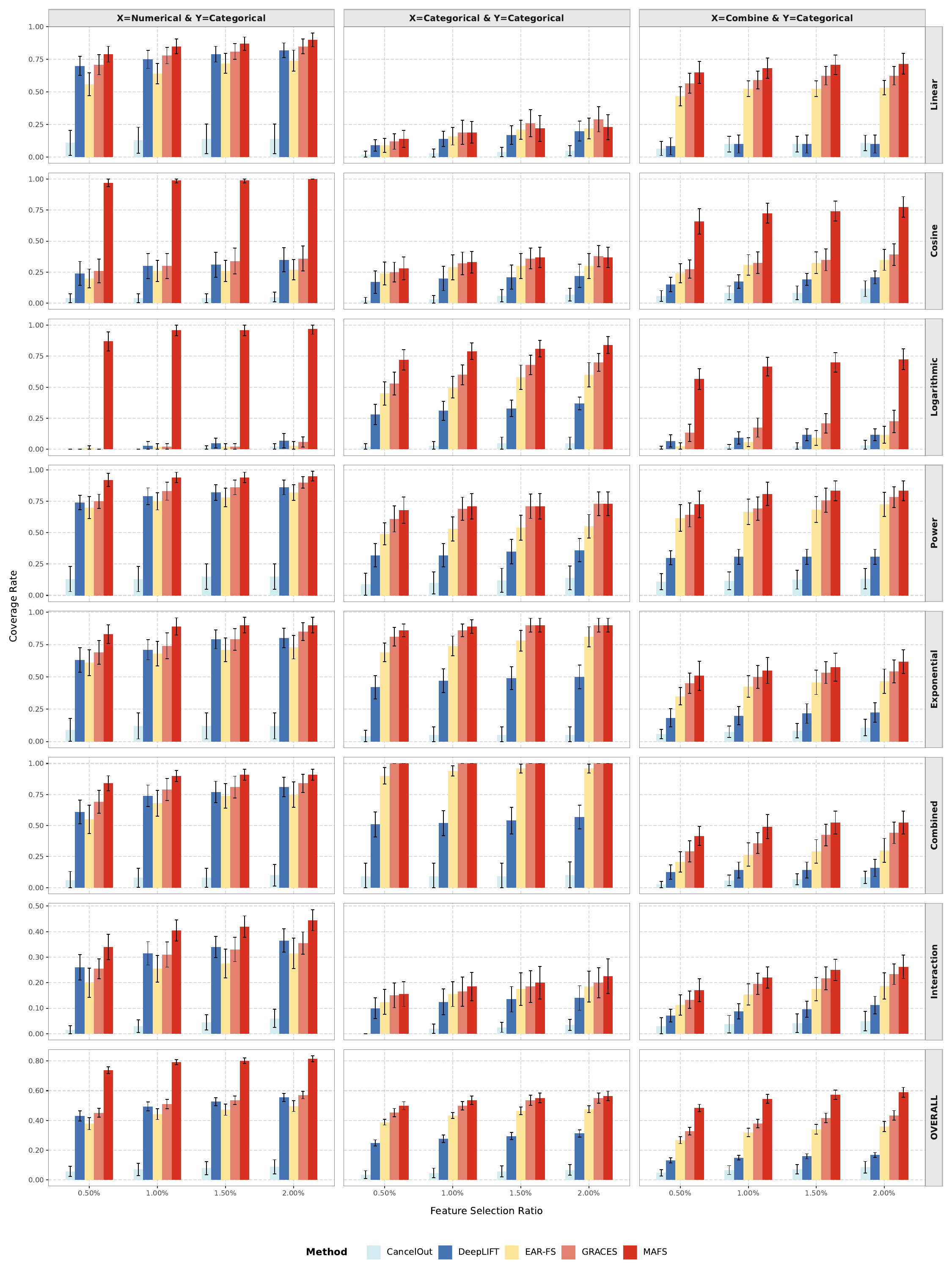}
    \caption{\textbf{Supplementary result of relationship-specific feature selection for categorical outcomes.} Coverage rates under seven different feature–response relationships at selection ratios 0.5\%, 1\%, 1.5\% and 2\% in the high-dimensional setting ($n=2{,}000$, $p=25{,}000$) for categorical outcomes.}
    \label{fig:relationship_analysis_2k_25k_cate}
\end{figure}
\begin{figure}[htbp] 
    \centering
    \includegraphics[width=0.95\textwidth]{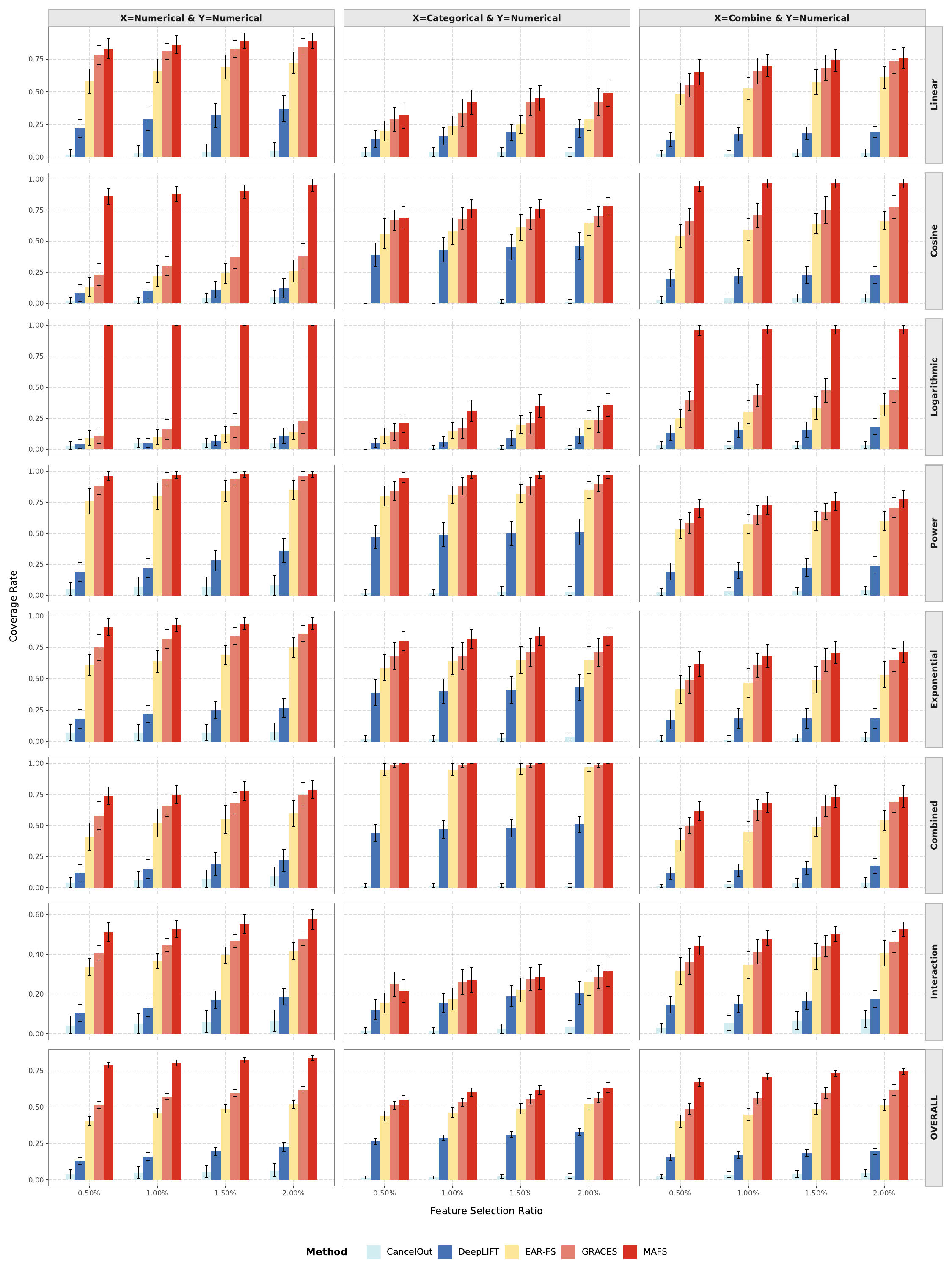}
    \caption{\textbf{Supplementary result of relationship-specific feature selection for continuous outcomes.} Coverage rates under seven different feature–response relationships at selection ratios 0.5\%, 1\%, 1.5\% and 2\% in the high-dimensional setting ($n=2{,}000$, $p=50{,}000$) for continuous outcomes.}
    \label{fig:relationship_analysis_2k_50k}
\end{figure}

\begin{figure}[htbp] 
    \centering
    \includegraphics[width=0.95\textwidth]{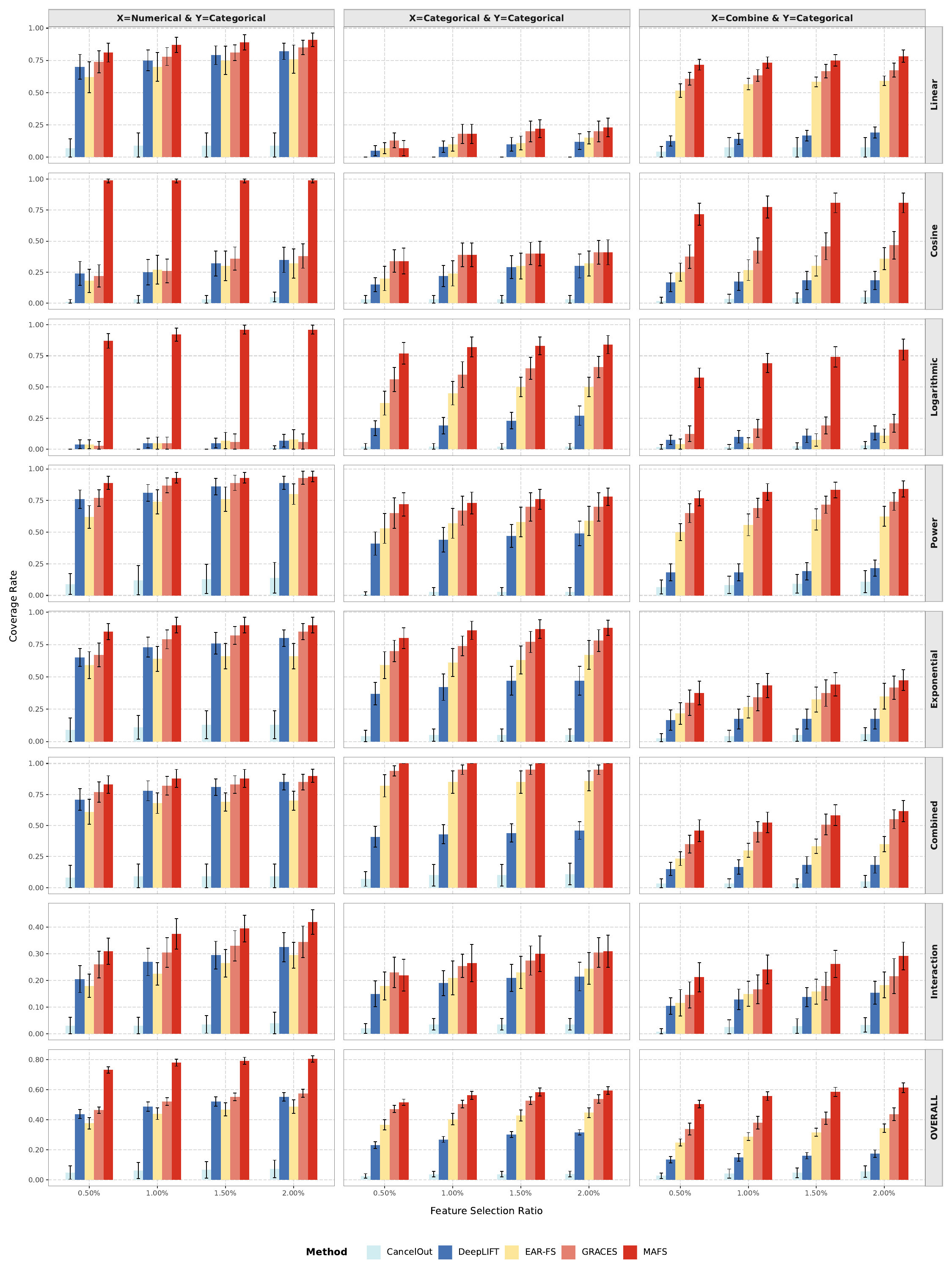}
    \caption{\textbf{Supplementary result of relationship-specific feature selection for categorical outcomes.} Coverage rates under seven different feature–response relationships at selection ratios 0.5\%, 1\%, 1.5\% and 2\% in the high-dimensional setting ($n=2{,}000$, $p=50{,}000$) for categorical outcomes.}
    \label{fig:relationship_analysis_2k_50k_cate}
\end{figure}

\newpage
\subsection*{Simulation 2 Supplement Results}
\begin{figure}[H]
    \centering
    \includegraphics[width=\textwidth]{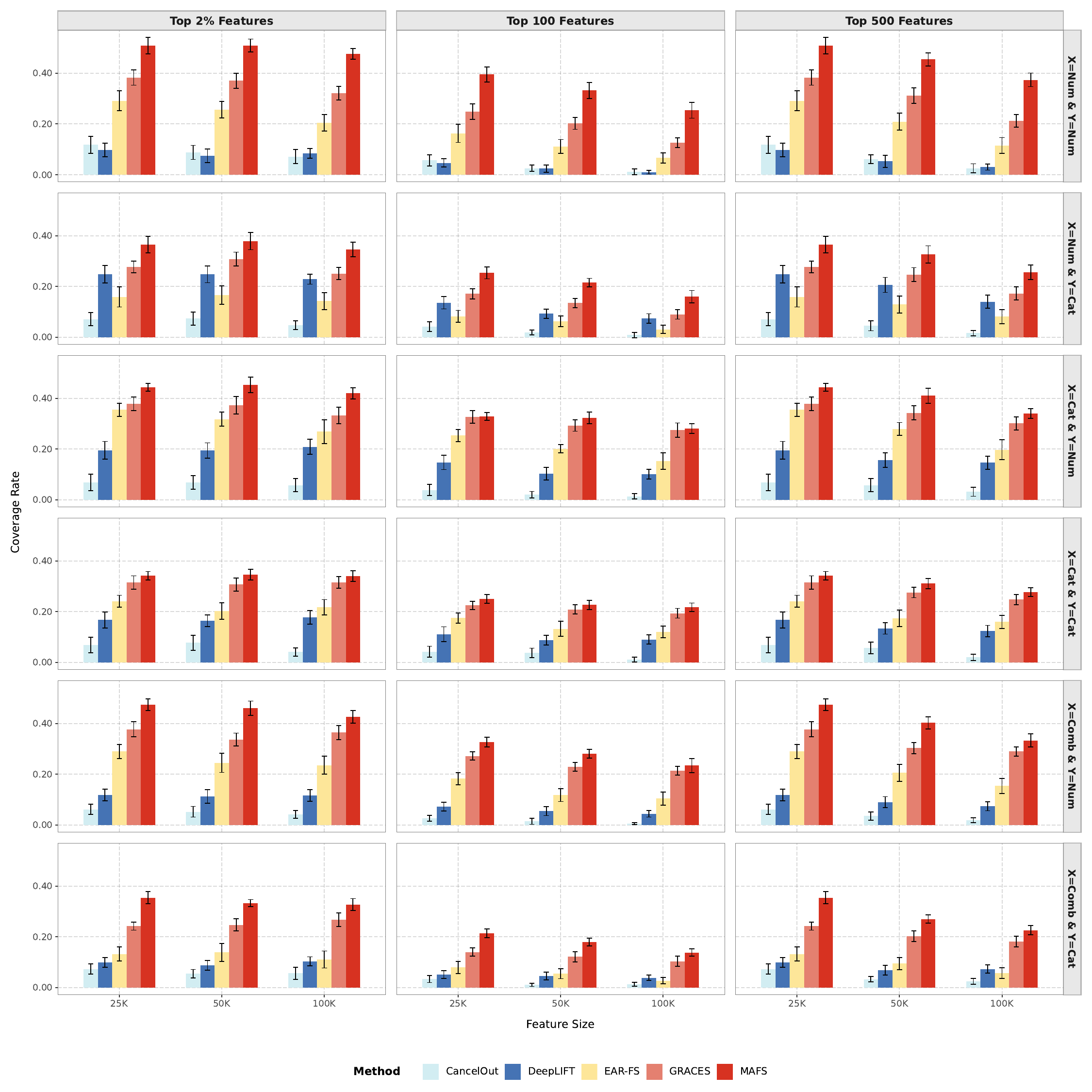}
    
    \caption{\textbf{Supplementary result performance across different dimensionalities.} Coverage rates under different input feature dimensions with moderate sample size ($n=500$). Each subfigure compares five feature selection methods across three feature dimensionalities (25K, 50K, and 100K features). Rows correspond to six data type combinations defined by feature types (continuous, categorical, and combined) and response types (continuous and binary). Columns correspond to three selection criteria: top 2\%, top 100, and top 500 features.}
    \label{fig:dimension_500}
\end{figure}


\newpage
\subsection*{Architecture Generalizability Validation}


\begin{figure}[H] 
    \centering
    \includegraphics[width=0.95\textwidth]{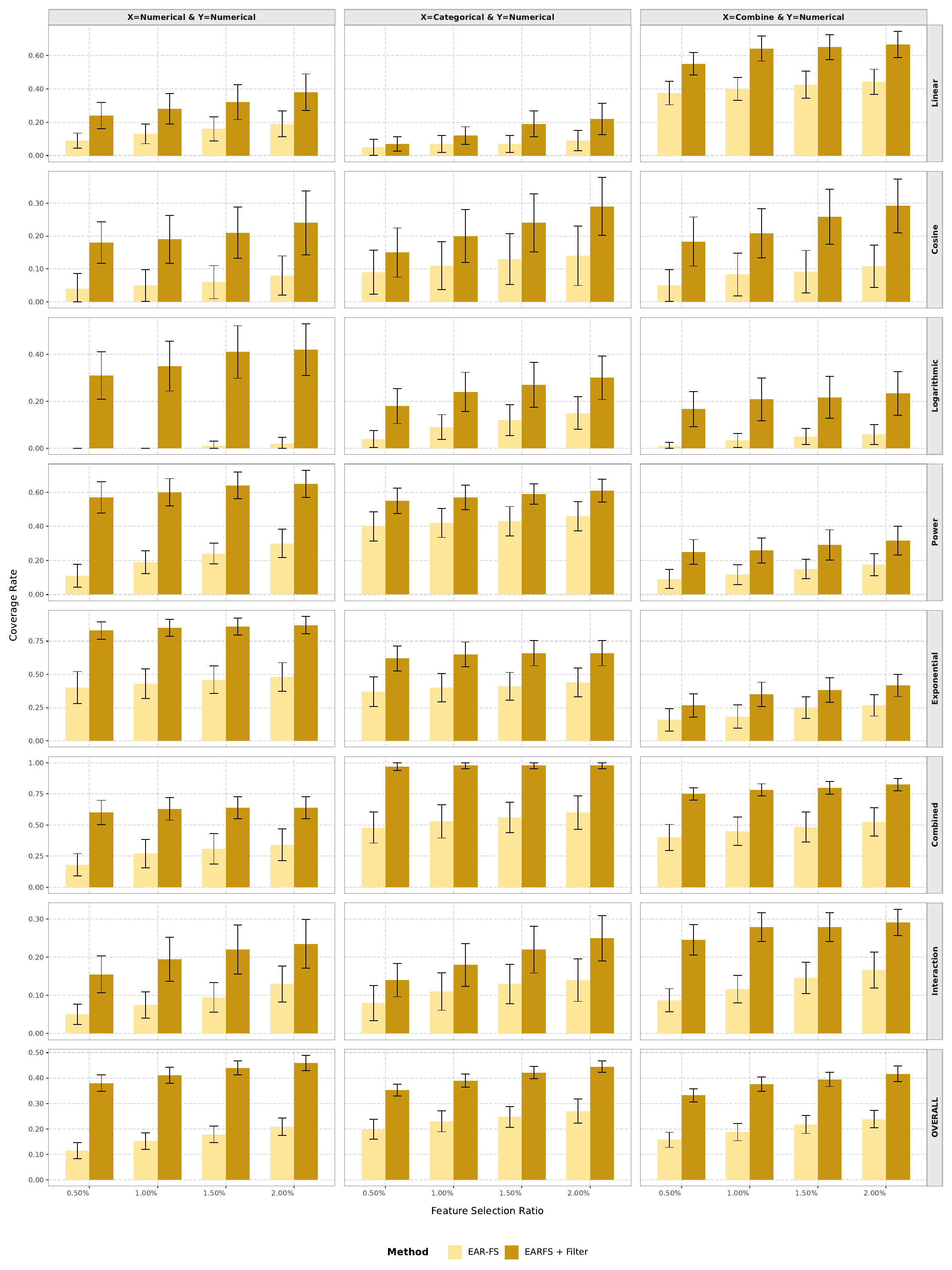}
        \caption{\textbf{Supplementary results on architecture generalizability for relationship-specific feature selection with continuous outcomes.} Coverage rates under seven different feature–response relationships at selection ratios 0.5\%, 1\%, 1.5\% and 2\% in the high-dimensional setting ($n=500$, $p=100{,}000$) for continuous outcomes.}
    \label{fig:relationship_analysis_500_part}
\end{figure}

\begin{figure}[H] 
    \centering
    \includegraphics[width=\textwidth]{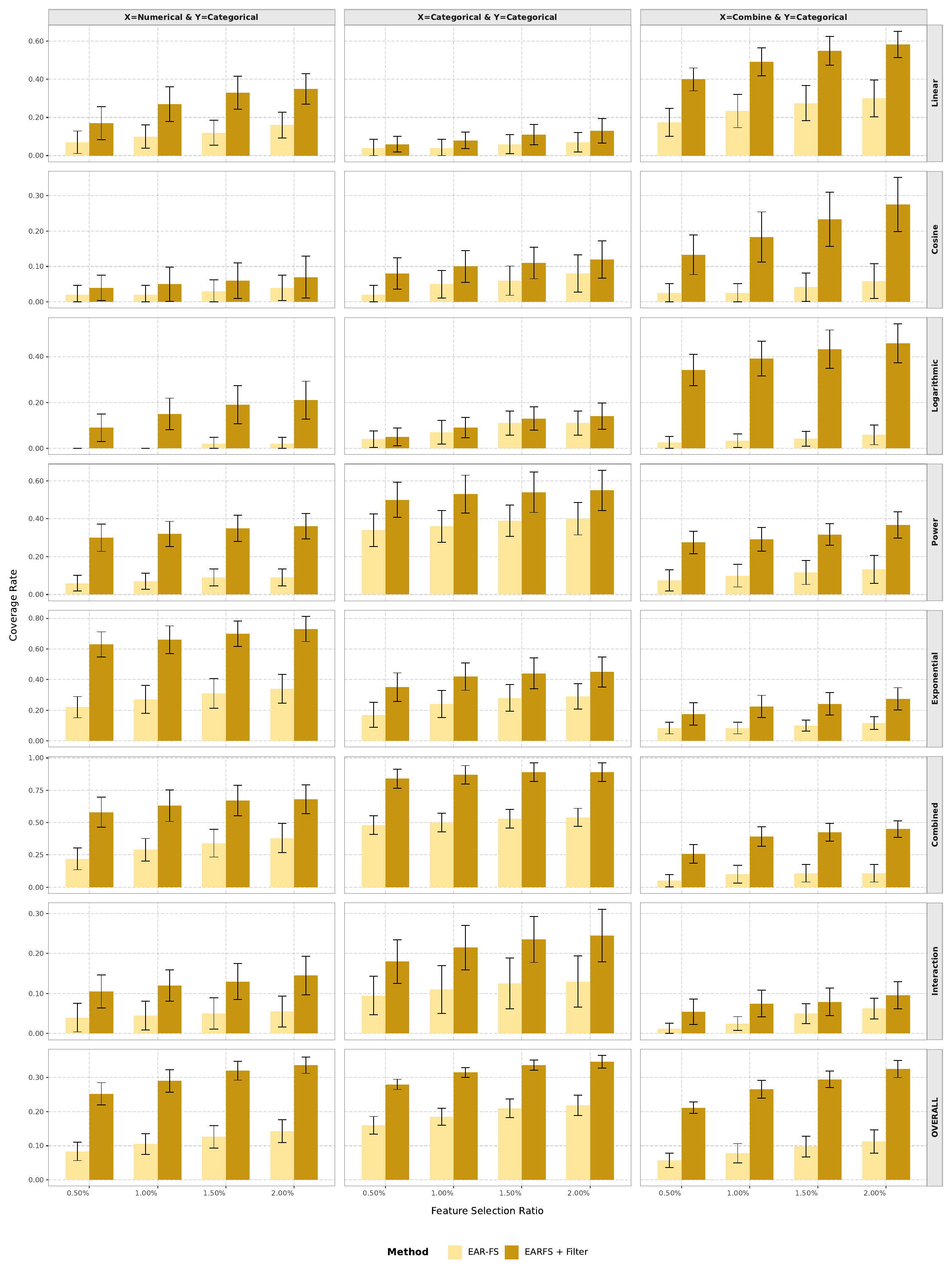}
    \caption{\textbf{Supplementary results on architecture generalizability for relationship-specific feature selection with categorical outcomes.} Coverage rates under seven different feature–response relationships at selection ratios 0.5\%, 1\%, 1.5\% and 2\% in the high-dimensional setting ($n=500$, $p=100{,}000$) for categorical outcomes.}
    \label{fig:relationship_analysis_500_cate_part}
\end{figure}


\begin{figure}[H] 
    \centering
    \includegraphics[width=\textwidth]{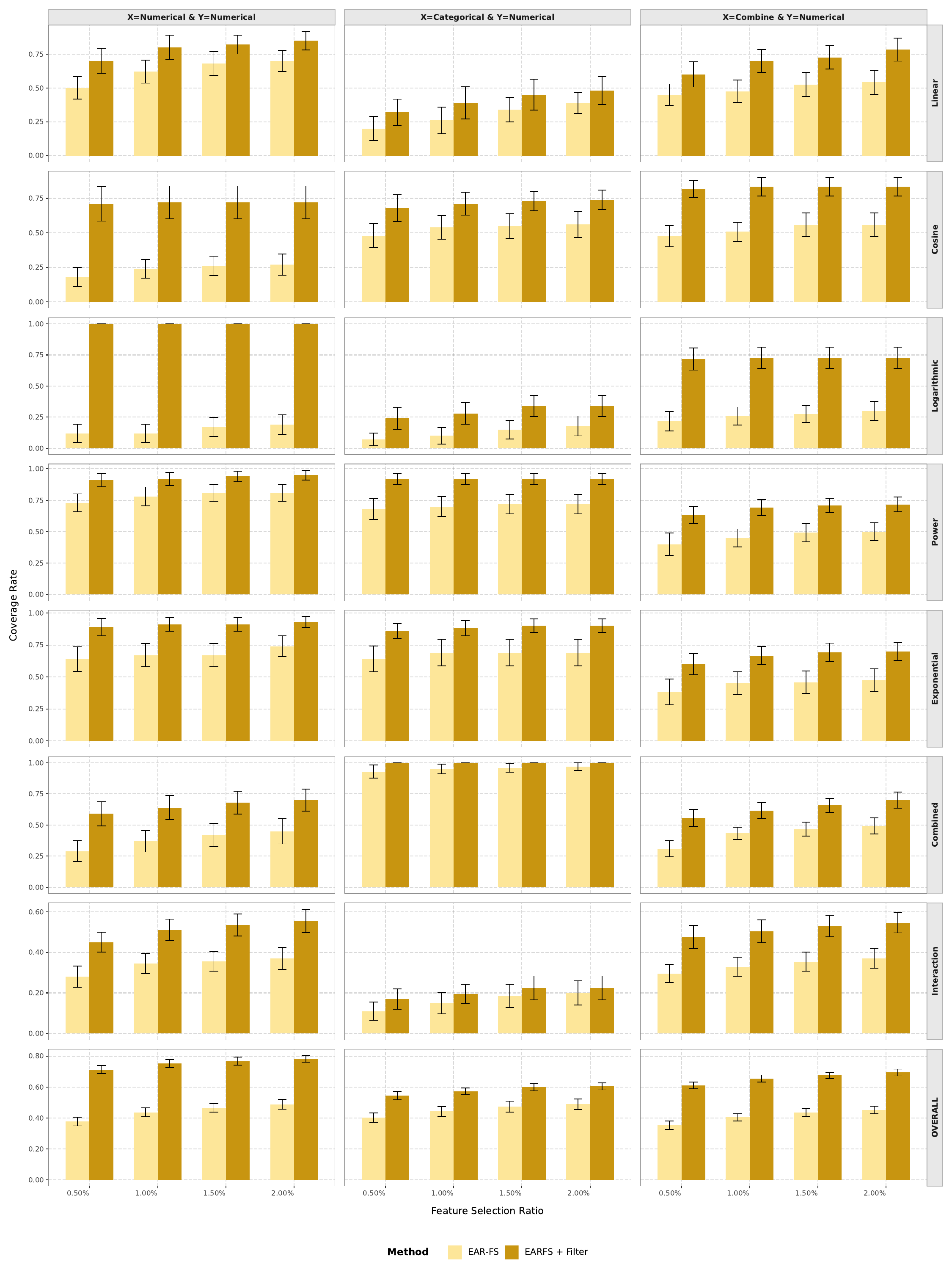}
        \caption{\textbf{Supplementary results on architecture generalizability for relationship-specific feature selection with continuous outcomes.} Coverage rates under seven different feature–response relationships at selection ratios 0.5\%, 1\%, 1.5\% and 2\% in the high-dimensional setting ($n=2{,}000$, $p=100{,}000$) for continuous outcomes.}
    \label{fig:relationship_analysis_2k_part}
\end{figure}

\begin{figure}[H] 
    \centering
    \includegraphics[width=\textwidth]{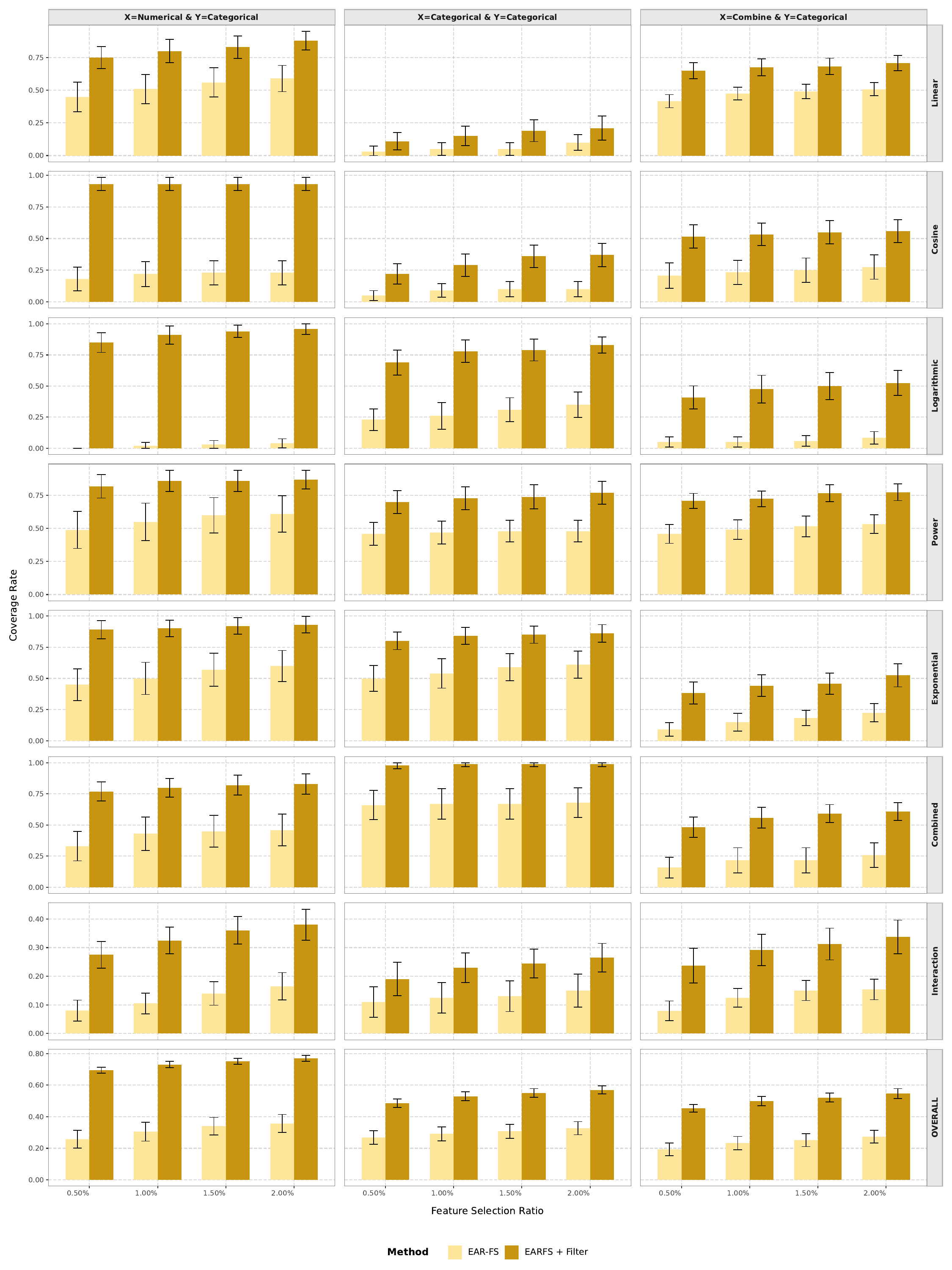}
    \caption{\textbf{Supplementary results on architecture generalizability for relationship-specific feature selection with categorical outcomes.} Coverage rates under seven different feature–response relationships at selection ratios 0.5\%, 1\%, 1.5\% and 2\% in the high-dimensional setting ($n=2{,}000$, $p=100{,}000$) for categorical outcomes.}
    \label{fig:relationship_analysis_2k_cate_part}
\end{figure}


\subsubsection*{Simualtion 2 Supplement Results}
\begin{figure}[H] 
    \centering
    \includegraphics[width=\textwidth]{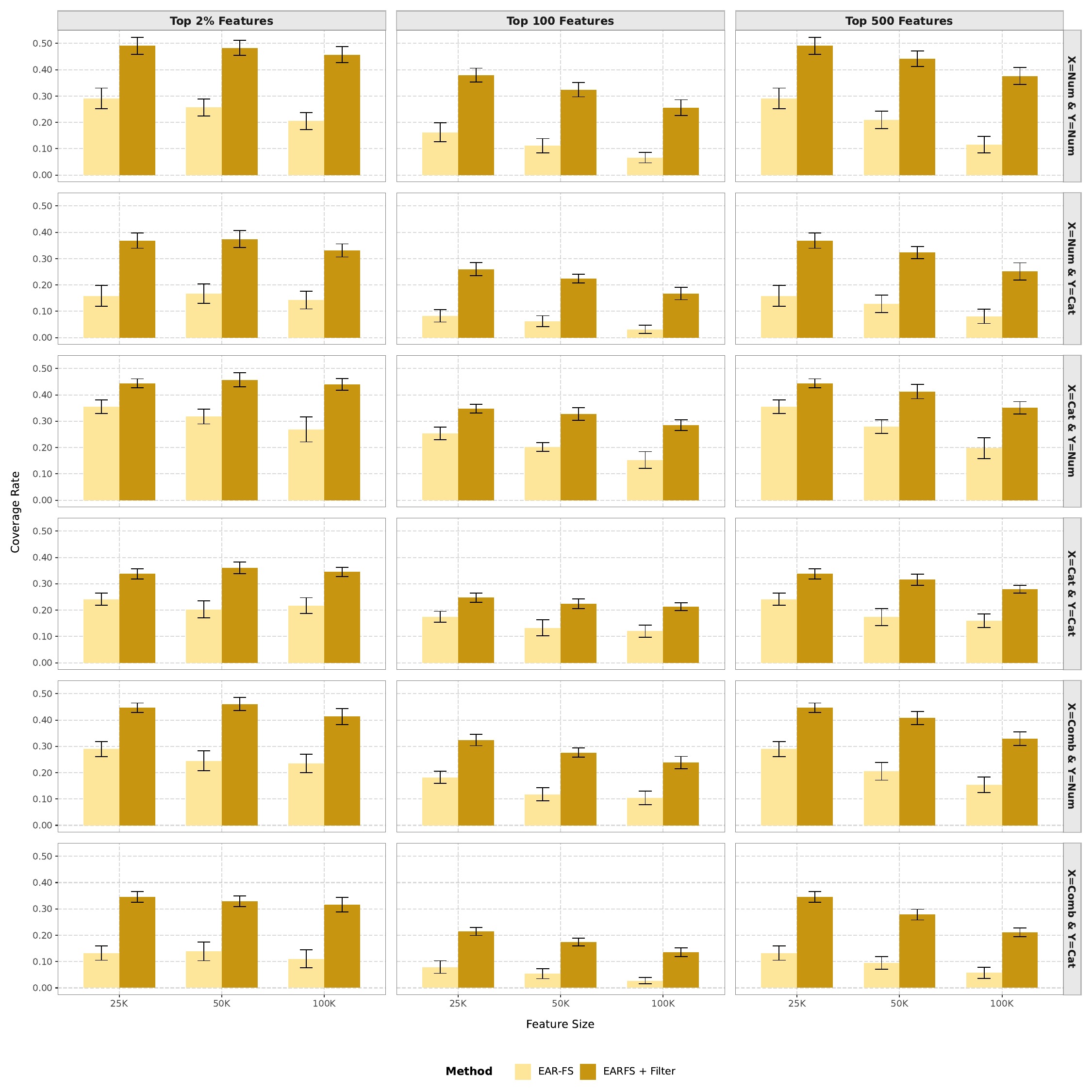}
    
    \caption{\textbf{Supplementary results on architecture generalizability across feature dimensionalities.} Feature selection performance under small-sample conditions ($n=500$). Each subfigure compares five feature selection methods across three feature dimensionalities (25K, 50K, and 100K features) for different selection proportions.  Columns correspond to feature type combinations (continuous, categorical, and combined), and rows correspond to response types (continuous and categorical). Columns correspond to three selection criteria: top 2\%, top 100, and top 500 features.}
    \label{fig:dimension_500_part}
\end{figure}

\begin{figure}[H] 
    \centering
    \includegraphics[width=\textwidth]{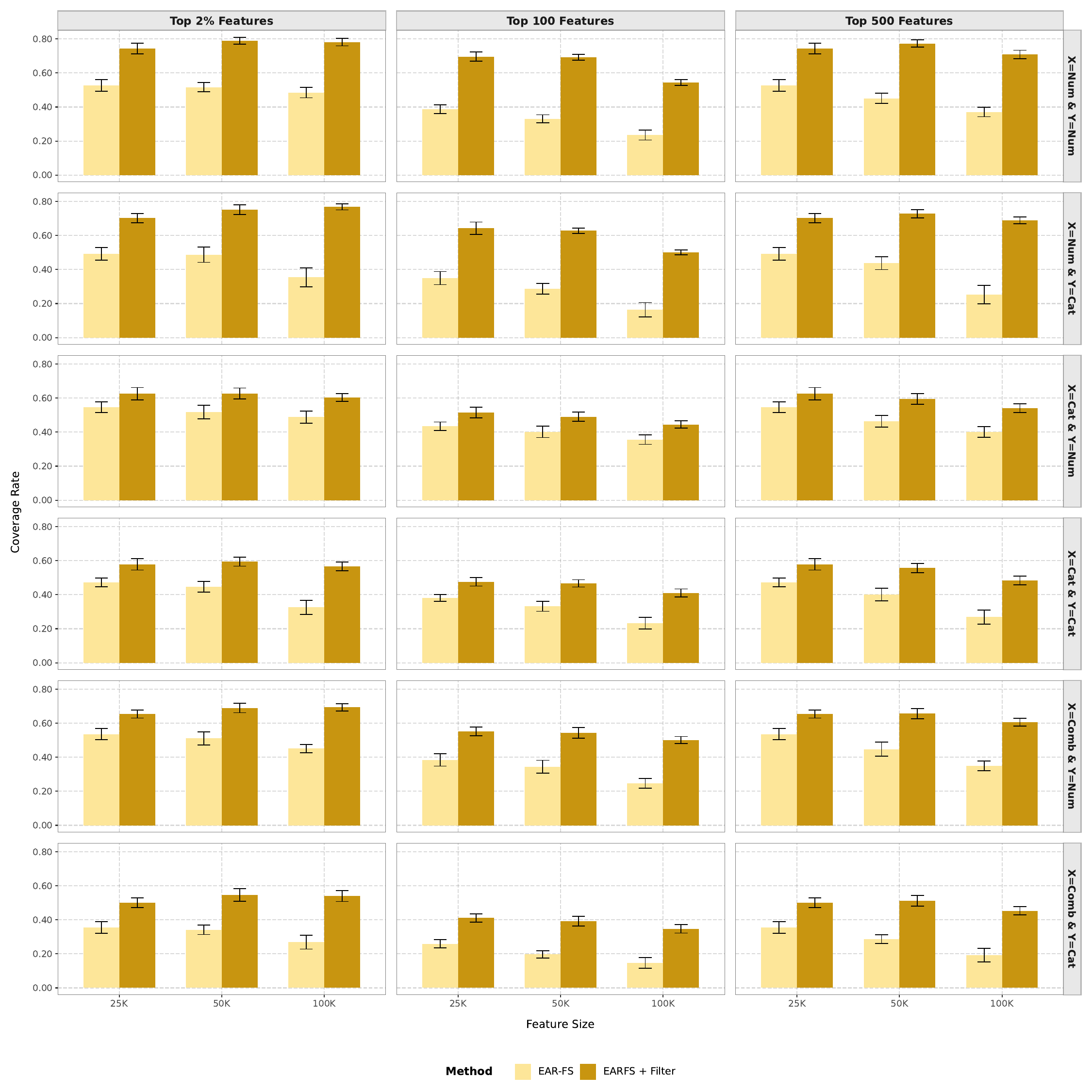}
    
    \caption{\textbf{Supplementary results on architecture generalizability across feature dimensionalities.} Feature selection performance under moderate-sample conditions ($n=2000$). Each subfigure compares five feature selection methods across three feature dimensionalities (25K, 50K, and 100K features). Rows correspond to six data type combinations defined by feature types (continuous, categorical, and combined) and response types (continuous and categorical). Columns correspond to three selection criteria: top 2\%, top 100, and top 500 features.}
    \label{fig:dimension_2k_part}
\end{figure}

\newpage

\begin{figure}[H] 
    \centering
    \includegraphics[width=1\textwidth]{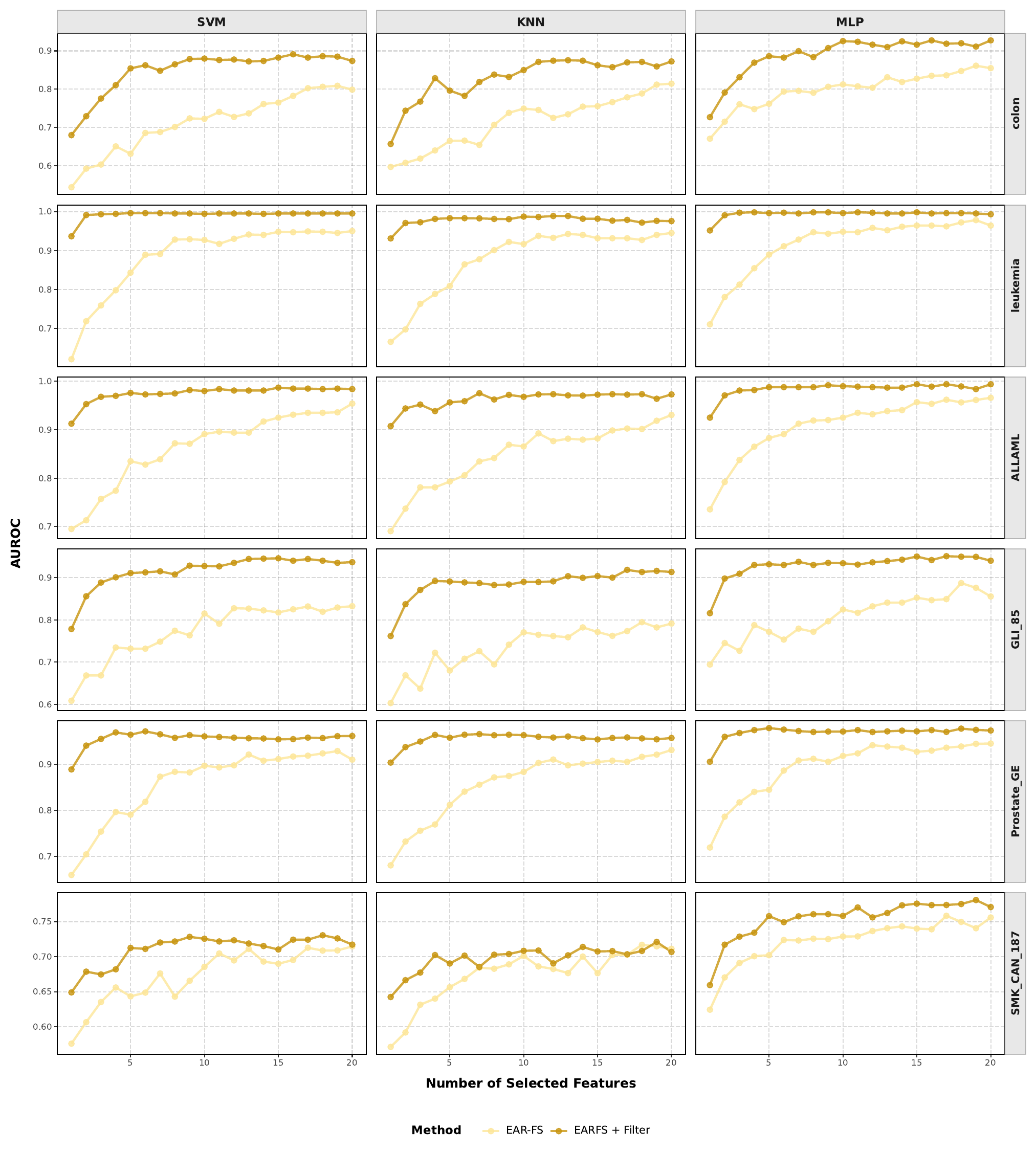}
    \caption{\textbf{Supplementary Results of Architecture Generalizability in Cancer Datasets.} Comparison of AUROC performance for feature selection methods across six cancer datasets. Each row corresponds to one dataset, and each column represents a different classifier: Support Vector Machine (left), K-Nearest Neighbors (middle), and Multi-Layer Perceptron (right).}
    \label{fig:auroc_part_classifiers}
\end{figure}

\newpage

\begin{figure}[H] 
    \centering
    \includegraphics[width=0.85\textwidth]{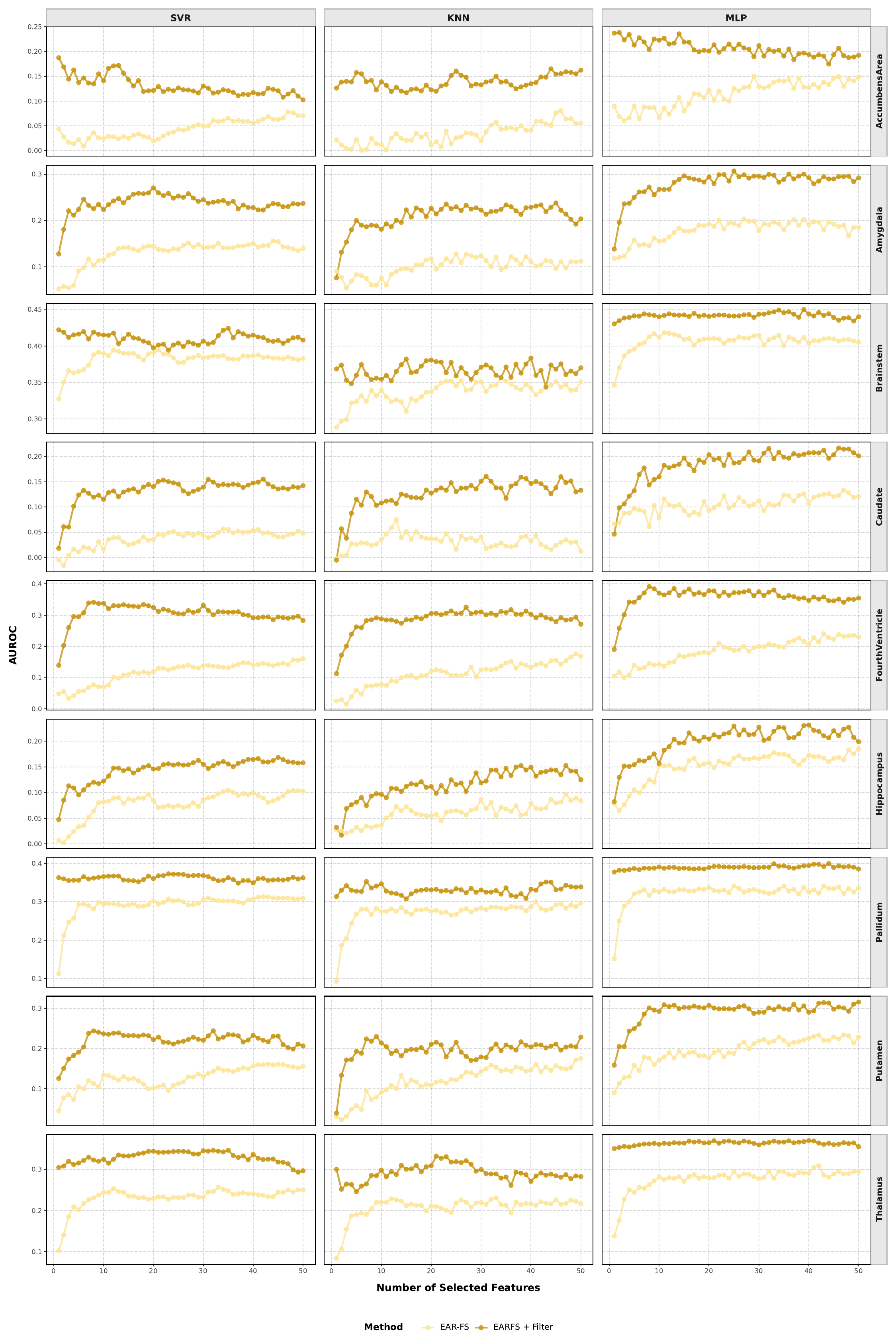}
    \caption{\textbf{Supplementary Results of Architecture Generalizability in ADNI Datasets.} Comparison of correlation performance for feature selection methods across the ADNI dataset. Each row corresponds to one brain region, and each column represents results obtained using a different regression model: Support Vector Regression (left), K-Nearest Neighbors Regression (middle), and Multi-Layer Perceptron Regression (right).}
    \label{fig:correlation_part}
\end{figure}